\pdfoutput=1

\documentclass[11pt]{article}

\usepackage[final]{acl}

\usepackage{times}
\usepackage{latexsym}
\usepackage{amsmath}
\usepackage{amssymb}
\usepackage{mathtools}
\usepackage{amsthm}
\usepackage{bbm}
\usepackage{makecell}
\usepackage{diagbox}
\usepackage{booktabs}
\usepackage{algorithm}
\usepackage{algorithmic}
\usepackage{hyperref}
\usepackage{subfigure}
\usepackage{tcolorbox}
\usepackage{colortbl}
\usepackage{color}
\definecolor{comment}{HTML}{A51C36}
\usepackage{wrapfig}
\usepackage{multirow}
\tcbuselibrary{breakable}
\newtcolorbox[auto counter, number freestyle={\noexpand\arabic{\tcbcounter}}]{mycolorbox}[3][]{%
    fonttitle=\bfseries,
    title=#2: #3,
    #1
}
\DeclareMathOperator*{\argmax}{arg\,max}

\usepackage[T1]{fontenc}

\usepackage[utf8]{inputenc}

\usepackage{microtype}

\usepackage{inconsolata}

\usepackage{graphicx}

\title{Examining False Positives under Inference Scaling \\
for Mathematical Reasoning}

\author{
 \textbf{Yu Wang\textsuperscript{1,2}}\textsuperscript{\dag},
 \textbf{Nan Yang\textsuperscript{2}}\textsuperscript{$\star$},
 \textbf{Liang Wang\textsuperscript{2}},
 \textbf{Furu Wei\textsuperscript{2}},
 \textbf{Fuli Feng\textsuperscript{1}}\textsuperscript{$\star$}
\\
 \textsuperscript{1}University of Science and Technology of China,
 \textsuperscript{2}Microsoft Research Asia
\\
 \texttt{terencewang0809@gmail.com}
\\
 \texttt{\{nanya, wangliang, fuwei\}@microsoft.com}
\\
 \texttt{fengfl@ustc.edu.cn}
\\
}

\makeatletter
\newenvironment{breakablealgorithm}
  {
   \begin{center}
     \refstepcounter{algorithm}
     \hrule height.8pt depth0pt \kern2pt
     \renewcommand{\caption}[2][\relax]{
       {\raggedright\textbf{Algorithm~\thealgorithm} ##2\par}%
       \ifx\relax##1\relax
         \addcontentsline{loa}{algorithm}{\protect\numberline{\thealgorithm}##2}%
       \else
         \addcontentsline{loa}{algorithm}{\protect\numberline{\thealgorithm}##1}%
       \fi
       \kern2pt\hrule\kern2pt
     }
  }{
     \kern2pt\hrule\relax 
   \end{center}
  }
\makeatother

\begin{document}
\maketitle

\let\thefootnote\relax\footnotemark\footnotetext{\textsuperscript{\dag} Work done during Yu’s internship at MSR Asia.}
\let\thefootnote\relax\footnotemark\footnotetext{\textsuperscript{$\star$} Corresponding authors}

\begin{abstract}
Recent advancements in language models have led to significant improvements in mathematical reasoning across various benchmarks. However, most of these benchmarks rely on automatic evaluation methods that only compare final answers using heuristics, without verifying the underlying reasoning steps. This limitation results in false positive solutions, where models may produce correct final answers but with flawed deduction paths. In this paper, we systematically examine the prevalence of false positive solutions in mathematical problem solving for language models. We analyze the characteristics and extent of this issue across different open-source models, datasets of varying difficulty levels, and decoding strategies. Specifically, we explore how ``false positives'' influence the inference time scaling behavior of language models. Our experimental results reveal that: (1) false positive solutions persist across different models, datasets, and decoding methods, (2) sampling-based inference time scaling methods do not alleviate the problem, and (3) the pass@N evaluation metric is more susceptible to ``false positives'', suggesting a significantly lower scaling ceiling than what automatic evaluations indicate. Additionally, we analyze specific instances of ``false positives'' and discuss potential limitations in self-improvement techniques and synthetic data generation under such conditions. Our data and code are publicly available at \href{https://github.com/Wloner0809/False-Positives-in-Math}{https://github.com/Wloner0809/False-Positives-in-Math}.
\end{abstract}

\section{Introduction}

Recent developments in language models, including improvements in inference time scaling and self-improvement techniques, have significantly enhanced performance in mathematical reasoning tasks \cite{snell2024scaling, shao2024deepseekmath}. However, many of these mathematical benchmarks rely on automatic evaluation methods that compare only the final answers generated by the models to reference answers, often using heuristic approaches. These methods do not guarantee the correctness of the reasoning steps taken to arrive at the final answer, raising concerns about the reliability of the evaluation metrics.

In this paper, we systematically investigate the prevalence and characteristics of ``false positives'' in mathematical problem-solving tasks. A ``false positive'' arises when the final answer is correct, but the solution process contains errors or lacks logical validity. We aim to provide a comprehensive analysis of how often ``false positives'' occur, and how they affect model performance across different open-source models, varying levels of difficulty in mathematical datasets, and diverse decoding strategies. Specifically, we select open-source models, including LLaMA (8B and 70B) \cite{dubey2024llama} and math-specialized models such as Qwen-Math (7B and 70B) \cite{yang2024qwen2}, to generate solutions. We test these models on three popular benchmarks—MATH \cite{hendrycks2021measuring}, AIME \cite{aime}, and OmniMATH \cite{gao2024omni}—which vary in difficulty. Additionally, we explore several sampling methods (e.g., repeated sampling, reward-guided beam search, and tree search) to assess how ``false positives'' influence inference time scaling. Both automatic and manual evaluation methods are employed to identify ``false positives''.

Our findings reveal several key insights. First, ``false positives'' are widespread across a variety of language models, datasets, and decoding methods, suggesting that this issue is not confined to any specific model or evaluation setup, but is a pervasive challenge in mathematical reasoning. Second, we explore whether different inference strategies, such as sampling-based methods, could mitigate the occurrence of ``false positives''. Our results show that these strategies do not significantly reduce the frequency of ``false positives'', indicating that the underlying reasoning flaws are not easily resolved by simply increasing the inference budget. Third, we observe that the pass@N metric is more susceptible to ``false positives'' than other metrics in automatic evaluations, which suggests that the ceiling for inference scaling may be substantially lower than what automatic evaluations indicate.

Additionally, we conduct a preliminary investigation into how rule-based GRPO \cite{guo2025deepseek} influences ``false positives'', as well as how the ``false positive'' phenomenon manifests in recent Long-CoT models. We analyze specific instances of ``false positives'', identifying the types of reasoning errors that lead to flawed solutions. We also discuss the implications of these findings for self-improvement techniques and synthetic mathematical data. In particular, we argue that ``false positives'' may limit the effectiveness of self-improvement methods, as models may appear to be learning correct reasoning patterns while, in reality, they are merely providing correct answers based on flawed deduction processes.

Ultimately, this paper aims to offer a more nuanced understanding of the challenges language models face in mathematical reasoning and to advocate for more rigorous evaluation practices that go beyond mere answer correctness.

\section{Related Work}
The ``false positive'' problem we investigate arises primarily due to the evaluation methods employed for assessing LLM performance on mathematical tasks. Many existing approaches focus solely on comparing the final answers to the ground truth. These evaluation strategies are efficient, inexpensive, and fully automated; however, they fail to account for the intermediate reasoning steps involved in generating the solution. Moreover, they are not applicable to tasks such as mathematical proofs, which do not have a single final answer. To overcome these limitations, some studies leverage powerful LLMs to compare the reasoning steps in generated solutions with reference answers or directly identify step errors within the reasoning path, in an attempt to evaluate the validity of mathematical reasoning \cite{he2023socreval, tyen2023llms, hao2024llm}. The effectiveness of this approach is heavily reliant on the capabilities of the LLM used, and it remains uncertain how reliably LLMs can detect reasoning flaws produced by strong LLMs themselves. Alternatively, other research has explored the use of formal proof systems for mathematical reasoning. Benchmarks such as MiniF2F \cite{zheng2021minif2f} and ProofNet \cite{azerbayev2023proofnet} utilize formal languages like Lean \cite{moura2021lean} to specify math problems, and LLMs are tasked with generating formal proofs, which can be rigorously checked by the formal system. While formal proofs inherently avoid the ``false positive'' issue present in natural language solutions, the translation of informal mathematical statements into formal systems remains a significant challenge, limiting the broader applicability of this approach.

Previous studies, such as \citet{hao2024llm} and \citet{zheng2024processbench}, have also highlighted the presence of ``false positives'' in LLM-generated mathematical solutions. A significant line of research focuses on improving the accuracy of reasoning steps through process supervision \cite{lightman2023let, setlur2024rewarding, luo2024improve}. For instance, \citet{lightman2023let} demonstrated that training on explicitly annotated flaws in reasoning paths could enhance the performance of reward models, leading to improved accuracy on mathematical benchmarks. In addition, studies like \citet{golovneva2022roscoe}, \citet{prasad2023receval} and \citet{xia2024evaluating} have proposed filtering and rescoring strategies, as well as novel metrics, to identify erroneous reasoning steps and mitigate the ``false positive'' problem. While \citet{snell2024scaling} investigated the inference time scaling of LLMs on mathematical problems, their work did not consider the impact of ``false positives''. Moreover, \citet{stroebl2024inference} studied how the ``false positive'' affects inference scaling in the coding domain, showing that flawed verifiers lead to a decrease in true accuracy as more computational resources are allocated, due to the growing rate of ``false positives''.

\section{Evaluation Methodology}
\label{section eval}

In mathematical evaluations, two primary assessment methods are commonly employed: automatic evaluation and manual evaluation. Automatic evaluation includes rule-based assessment and the use of powerful LLMs for evaluation. Currently, most benchmarks for mathematical models typically rely on rule-based automatic evaluation \cite{yang2024qwen2, shao2024deepseekmath}. This approach utilizes predefined heuristic rules to evaluate the correctness of a model's output by comparing its final answer to the ground truth. While this method is straightforward and easy to implement, it has notable limitations. Specifically, it fails to effectively assess the correctness and logical coherence of intermediate reasoning steps, leading to the phenomenon of ``false positives''. The detection of such ``false positives'' can be conducted either through model-based methods or human evaluation.

\subsection{Model Detection of False Positives}
To assess the ability of current models to detect errors in intermediate reasoning steps, we can utilize $\mathcal{M}(\text{True}\ or\ \text{False}\ |\ \mathcal{P},x,y)$, where $\mathcal{M}$ denotes the model used for error detection, $x$ and $y$ represent the question and the model-generated response, respectively, and $\mathcal{P}$ refers to the prompt utilized (see Appendix \ref{model detection prompt} for details). Although the costs of employing the model for error detection are relatively low, its effectiveness remains limited. We present a comprehensive analysis and discussion in Section \ref{model detection experiment}.

\subsection{Human Detection of False Positives}
\label{human detection method}
Due to the limited capability of existing models to identify errors in reasoning steps, we introduce manual evaluation as a complementary approach to better understand the occurrence of ``false positives''. Human evaluation involves a meticulous, step-by-step review of the model's responses by human annotators, ensuring not only the correctness of the final answer but also the logical coherence and mathematical validity of the intermediate steps. While more resource-intensive, this method significantly improves the accuracy and comprehensiveness of the evaluation, providing deeper insights into the model's reasoning processes.

In human evaluation, annotators classify a model's response as a ``false positive'' if it exhibits any of the following errors, despite the final answer being correct:

\begin{enumerate}
    \item \textbf{Jump in Reasoning}: This occurs when essential logical steps or intermediate calculations are omitted, resulting in a direct leap to the final answer without adequate justification. Such omissions undermine the validity of the solution, even if the answer itself is correct.
    \item \textbf{Logical Error}: This category encompasses errors such as the misapplication of theorems or rules, reliance on unjustified assumptions, contradictory reasoning, and the incorporation of conditions absent from the problem statement.
    \item \textbf{Calculation Error}: Mistakes in arithmetic or algebraic computations, while potentially offset by other errors, still reflect a lack of precision in the solution process.
    \item \textbf{Conceptual Error}: Misinterpretation of mathematical theorems, concepts, or the problem itself.
\end{enumerate}

Additionally, human annotators may disregard minor errors in the reasoning path that do not affect the final answer. Furthermore, if the model successfully identifies and corrects its own mistakes through self-correction or reflection, such reasoning paths are considered valid and are not labeled as ``false positives''. In Section \ref{results}, we primarily examine the false positive phenomenon in Short-CoT models and also discuss its manifestation in Long-CoT models. For Long-CoT models, we focus only on the <answer> part of the model output.

\begin{figure*}[t]
\centering  
{
\includegraphics[width=12cm,height = 4.5cm]{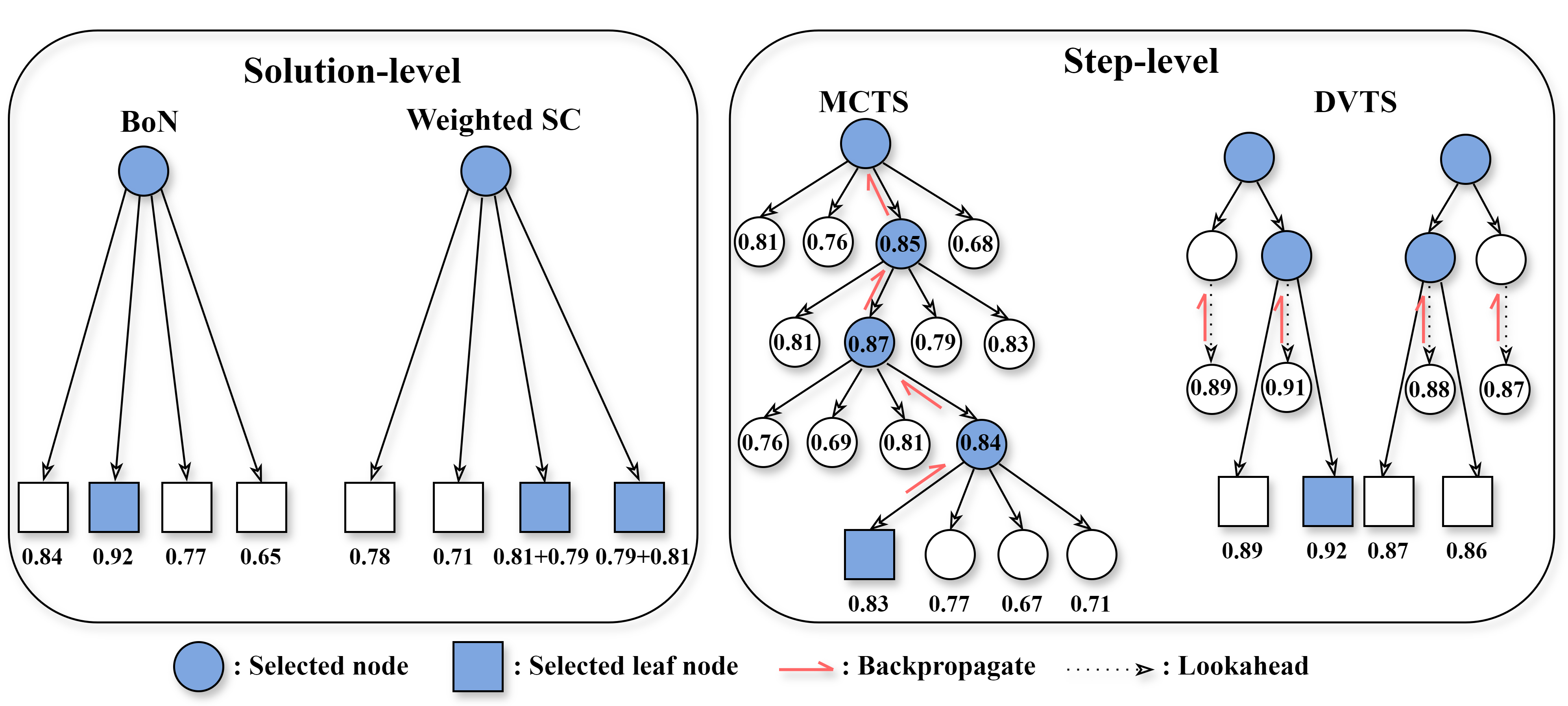}
}
\caption{An overview of the inference scaling methods employed in this study. The numbers in the figure indicate potential PUCT scores for MCTS and possible reward examples from the Process Reward Model (PRM) or Outcome Reward Model (ORM) for other methods. For MCTS, the figure depicts its first iteration.}
\label{method}
\end{figure*}

\section{Inference Scaling Methods in the Presence of False Positives}
Recent studies have demonstrated that allocating additional computational resources to the inference phase can significantly enhance model performance in mathematical tasks \cite{snell2024scaling, wu2024empirical}. However, these studies typically rely on rule-based evaluation methods, which may result in ``false positives'', as discussed in Section \ref{section eval}. To investigate whether ``false positives'' also manifest in inference scaling, this section offers a comprehensive overview of the inference scaling methods utilized in our study.

We classify current sampling-based inference scaling methods into two categories: \textbf{solution-level inference scaling} and \textbf{step-level inference scaling}. Solution-level inference scaling refers to the approach where LLMs generate a complete reasoning path in a single sampling process, with the optimal solution selected using a reward model or heuristic methods. In contrast, step-level inference scaling derives one reasoning step at a time, typically guided by a reward model or heuristic values. Figure \ref{method} provides a visual representation of these methods, and the following sections offer a detailed explanation of each approach.
\subsection{Solution-Level Inference Scaling}
Currently, the most widely used solution-level inference scaling methods are Best-of-N \cite{BoN1, BoN2}, Self-Consistency \cite{SC}, Weighted Self-Consistency \cite{WeightedSC}. Let $\mathcal{Y}$ signify the output space of large language models, $\mathcal{A}$ correspond to the answer space, where answers are extracted from the model outputs, and $v:\mathcal{Y}\xrightarrow{}\mathbb{R}$ represent the score function.
\begin{enumerate}
    \item \textbf{Best-of-N}: Best-of-N is a simple yet effective reranking algorithm \cite{welleck2024decoding}. It begins by generating $N$ candidate solutions, and subsequently selects the one with the highest score assigned by the score function. Best-of-N can be defined as: $y^\star=\argmax_{y_i\in\mathcal{Y}|i\in\{1,\cdots,N\}}v(y_i)$
    \item \textbf{Self-Consistency}: Self-Consistency is a transformation algorithm \cite{welleck2024decoding}, leveraging the idea that correct reasoning processes, though diverse, often converge on the same answer. This method first samples $N$ candidate reasoning paths and then determines the final answer by selecting the one that appears most frequently. Self Consistency is formally expressed as: $a^\star=\argmax_{a\in\mathcal{A}}\sum\limits_{i=1}^N\mathbbm{1}(a=a_i)$
    \item \textbf{Weighted Self-Consistency}: Weighted Self-Consistency extends Self-Consistency by incorporating the scores provided by the reward model to weigh candidate solutions. The optimal answer is chosen using the following formula: $a^\star=\argmax_{a\in\mathcal{A}}\sum\limits_{i=1}^N v(y_i)\mathbbm{1}(a=a_i)$
\end{enumerate}

In subsequent experiments, we employ Best-of-N and Weighted Self-Consistency instead of Self-Consistency. While both Weighted Self-Consistency and Self-Consistency are based on the principle of consistency, Weighted Self-Consistency exhibits superior performance \cite{snell2024scaling}. Within the Weighted Self-Consistency algorithm, we first use it to select the candidate final answer. To further detect ``false positives'', the solution with the highest reward among these candidates is selected as the final target for evaluation.

\subsection{Step-Level Inference Scaling}
\subsubsection{Diverse Verifier Tree Search}
\textbf{Diverse Verifier Tree Search} (DVTS, \cite{beeching2024scalingtesttimecompute}) is an extension of step-level beam search \cite{welleck2022naturalprover, yao2024tree} that divides initial beams into independent subtrees. The search process in DVTS is guided by a Process Reward Model (PRM). Additionally,  DVTS incorporates lookahead steps to enhance the accuracy of PRM's value estimation at each step of the search process.

The detailed specifics of DVTS can be found in Algorithm \ref{dvts}.

\subsubsection{Monte Carlo Tree Search}
\textbf{Monte Carlo Tree Search} (MCTS, \cite{browne2012survey}) is a tree search algorithm designed to balance exploration and exploitation effectively. In this work, We utilize the Vanilla MCTS implementation from \citet{wang2024openr}, which comprises four main steps: selection, expansion, evaluation, and backpropagation. During the selection stage, Vanilla MCTS employs a variant of the PUCT algorithm \cite{silver2016mastering} to choose child nodes. In the evaluation stage, it leverages PRM to compute state values. Each iteration of the algorithm continues until a complete reasoning path is obtained.

For further details on Vanilla MCTS, see Algorithm \ref{mcts}.

\section{Experiments}
\subsection{Experimental Setup}
\textbf{Benchmarks.} To validate the proposed phenomenon, we conduct experiments on three mathematical benchmarks: MATH \cite{hendrycks2021measuring}, AIME \cite{aime}, Omni-MATH \cite{gao2024omni}. MATH comprises problems collected from high school math competitions. Following \citet{lightman2023let}, we use MATH500 as our test set. AIME includes questions from AIME\{22, 23, 24\}, totaling 90 problems. Omni-MATH is a highly challenging benchmark designed for Olympiad-level mathematical reasoning, and we utilize Omni-MATH-Rule \cite{gao2024omni}, a subset suitable for rule-based evaluation. For further convenience of manual evaluation, We randomly select 100 problems from MATH500 and 100 problems from Omni-MATH-Rule, which we refer to as MATH100 and Omni-MATH100 respectively.

\noindent\textbf{Policy Models.} We select open-source general and mathematical models as our base to investigate whether the proposed phenomenon is present in both types of models. Specifically, Llama-3.2-3B-Instruct and Llama-3.1-\{8B, 70B\}-Instruct \cite{dubey2024llama} are chosen to represent general-purpose models, while Qwen2.5-Math-\{1.5B, 7B, 72B\}-Instruct \cite{yang2024qwen2} serve as representatives of mathematical models.

\noindent\textbf{Reward Models.} We leverage both the Outcome Reward Model (ORM) and the Process Reward Model (PRM). For ORM, we employ Qwen2.5-Math-RM-72B \cite{yang2024qwen2} to identify the optimal model response in solution-level inference scaling methods. For PRM, we utilize Skywork-o1-Open-PRM-Qwen-2.5-7B \cite{skyworkopeno12024}, which is fine-tuned on Qwen2.5-Math-7B-Instruct, to guide DVTS or MCTS processes.

\noindent\textbf{Metrics.} We primarily employ three metrics to evaluate performance: automatic accuracy, false positive rate, and manual accuracy. \textit{Automatic accuracy} is computed using rule-based methods, following the implementation in Qwen2.5-Math \cite{yang2024qwen2}. \textit{False positive rate} is defined as the proportion of ``false positives'' among all responses deemed correct by the automatic evaluation. \textit{Manual accuracy} is determined through human evaluation and reflects the proportion of model responses that both match the ground truth and are free of ``false positives''. \\

\noindent Refer to Appendix \ref{implementation appendix} for additional parameter settings and implementation details.

\begin{figure*}[t]
\centering
\subfigure[Solution-Level Inference Scaling Methods]
{
\includegraphics[width=7.7cm,height = 3.2cm]{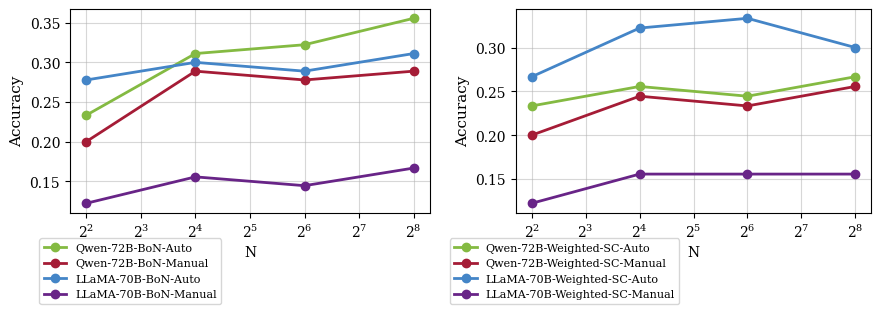}
}
\subfigure[Step-Level Inference Scaling Methods]
{
\includegraphics[width=7.7cm,height = 3.2cm]{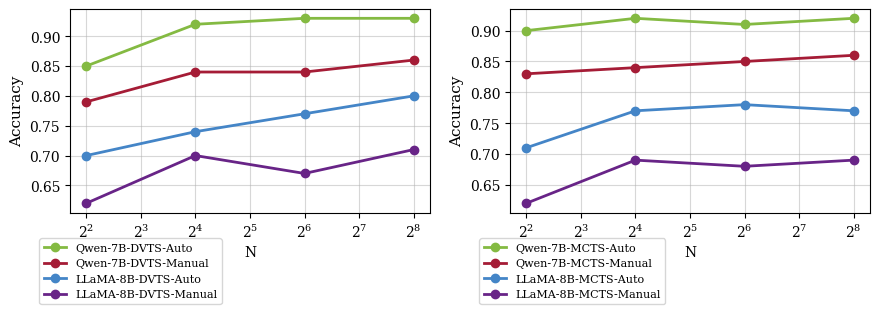}
}
\caption{\textbf{False Positives in Different Inference Scaling Methods}. For solution-level inference scaling methods, Llama-3.1-70B-Instruct and Qwen2.5-Math-72B-Instruct are used as policy models, with AIME as the evaluation dataset. Qwen2.5-Math-RM-72B is employed as the reward model. For step-level inference scaling methods, Llama-3.1-8B-Instruct and Qwen2.5-Math-7B-Instruct serve as policy models, evaluated on MATH100. Skywork-o1-Open-PRM-Qwen-2.5-7B is selected as the reward model. Responses from Llama-3.1-8B-Instruct are selected using Best-of-N, while responses from Qwen2.5-Math-7B-Instruct are chosen using Weighted Self-Consistency.}
\label{false positive}
\end{figure*}

\subsection{Model and Human Detection of False Positives}
\label{model detection experiment}
Prior to investigating the ``false positive'' phenomenon, we first analyze the differences between the capabilities of models and humans in detecting ``false positives''. To this end, we construct a comprehensive false positive detection benchmark that encompasses multiple models, diverse mathematical benchmarks, and various inference scaling methods (Further details regarding the benchmark are provided in Appendix \ref{false positive detection benchmark}). The $F_1$-score is utilized as the evaluation metric for this task. Let $A$ represent the set of ``false positives'' identified through manual evaluation, which serves as the gold standard, and $B$ denote the set of ``false positives'' identified by the model. Precision and recall are defined as follows: $$Precision=\frac{\#(A\cap B)}{\# B},\ Recall=\frac{\#(A\cap B)}{\# A}$$ Thus, the $F_1$-score is calculated as: $$F_1=\frac{2\times Precision\times Recall}{Precision+Recall}$$

\begin{figure}[t]
\centering  
{
\includegraphics[width=6cm,height = 4cm]{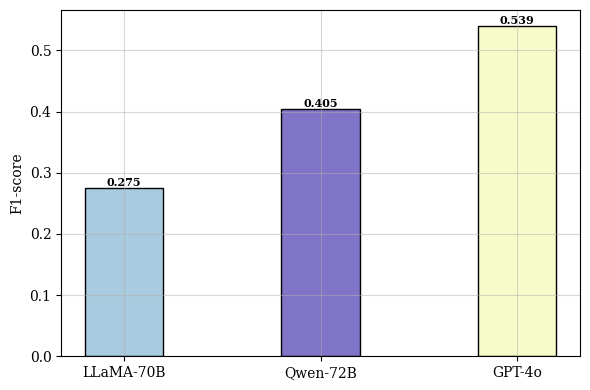}
}
\caption{\textbf{$F_1$-Scores Across Different False Positive Detection Models}.}
\label{model_check}
\end{figure}

Figure \ref{model_check} demonstrates that current open-source models, despite their overall strong performance, struggle to effectively detect errors in intermediate reasoning steps on the false positive detection benchmark we construct. Notably, even the powerful closed-source model, GPT-4o \cite{hurst2024gpt}, fails to achieve satisfactory performance in this regard.

\begin{figure}[t]
\centering  
{
\includegraphics[width=6cm,height = 4cm]{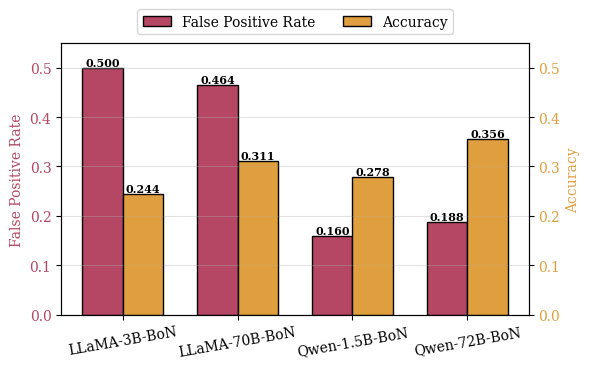}
}
\caption{\textbf{False Positive Rate and Automatic Accuracy in Different Models}. The performance of four policy models is assessed on AIME with the Best-of-N method, where $N$ is set to 256 and Qwen2.5-Math-RM-72B serves as the reward model.}
\label{different model}
\end{figure}

\subsection{Findings and Analysis}
\label{results}
\subsubsection{False Positives in Inference Scaling}
\label{false positives}
In this section, we investigate the accuracy of the inference scaling curve using solution-level and step-level inference scaling methods. Specifically, We analyze the relationship between automatic accuracy and manual accuracy across varying values of $N$ in different approaches. Additionally, we explore the impact of model types and benchmark difficulty on the false positive rate.

\noindent\textbf{False positives occur in both inference scaling methods.} As shown in Figure \ref{false positive}, both automatic accuracy and manual accuracy generally increase with $N$ across all methods. However, the gap between automatic accuracy and manual accuracy persists when $N$ takes different values, indicating a consistent presence of ``false positives''. This observation reveals that inference scaling curves are not as reliable as they might initially appear and that inference scaling does not effectively mitigate ``false positives''.

\noindent\textbf{General models exhibit higher false positive rates than mathematical models.} Figure \ref{different model} shows how false positive rate and accuracy vary across different model types. General models, which are not specialized in solving mathematical problems, exhibit significantly higher false positive rates than mathematical models on the relatively challenging AIME benchmark, regardless of whether their automatic accuracy is higher or lower. This suggests that the correct answers generated by inference scaling in general models are less reliable than those produced by mathematical models for difficult datasets. Additionally, it is observed that the smaller Qwen2.5-Math-1.5B-Instruct model yields a lower false positive rate compared to the larger Llama-3.1-70B-Instruct model. This is primarily due to the Llama-3.1-70B-Instruct model producing a greater number of reasoning jumps and logical errors. We discuss further in Section \ref{quantitative analysis}.

\begin{figure}[t]
\centering  
{
\includegraphics[width=6cm,height = 4cm]{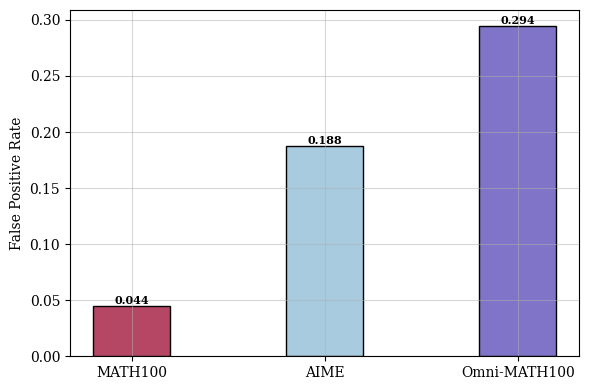}
}
\caption{\textbf{False Positive Rate in Different Datasets}. We employ the Best-of-N method to evaluate Qwen2.5-Math-72B-Instruct's performance on three datasets of varying difficulty levels, with Qwen2.5-Math-RM-72B as the reward model and $N$ set to 256.}
\label{different dataset}
\end{figure}

\begin{table}[t]
\caption{\textbf{Comparison of Solution Lengths Across Different Benchmarks and Solution Types (Same Setup as Figure \ref{different dataset})}. \textit{Type1: All Solutions, Type2: Final Answer Correct Solutions, Type3: False Positive Solutions}.}
\label{tab:solution_lengths_merged}
\resizebox{0.48\textwidth}{!}{
\begin{tabular}{@{}lcccc@{}}
\toprule
Benchmark & Sol. Type & Avg. Len. & Max. Len. & Min. Len. \\
\midrule
\multirow{3}{*}{MATH100} & \textit{Type1} & 571.92 & 1738 & 185 \\
& \textit{Type2} & 537.39 & 1279 & 193 \\
& \textit{Type3} & 875.75 & 1279 & 550 \\
\midrule
\multirow{3}{*}{AIME} & \textit{Type1} & 1116.59 & 4127 & 231 \\
& \textit{Type2} & 972.91 & 2217 & 485 \\
& \textit{Type3} & 1169.17 & 1780 & 588 \\
\midrule
\multirow{3}{*}{Omni-MATH100} & \textit{Type1} & 880.04 & 4180 & 197 \\
& \textit{Type2} & 701.67 & 1918 & 197 \\
& \textit{Type3} & 1019.33 & 1918 & 429 \\
\bottomrule
\end{tabular}
}
\end{table}

\noindent\textbf{False positive rates increase with benchmark difficulty.} The results in Figure \ref{different dataset} show that the false positive rate for Qwen2.5-Math-72B-Instruct rises as dataset difficulty increases. Notably, the false positive rates for more challenging datasets differ significantly from those for simpler ones. This finding highlights the tendency of inference scaling to produce ``false positives'' when tackling problems that exceed the model's inherent capabilities.

To gain deeper insight into why false positive rates increase with benchmark difficulty, we analyze the model's output length, as summarized in Table \ref{tab:solution_lengths_merged}. The table shows that, across all three benchmarks, the average length of \textit{False Positive Solutions} is greater than that of \textit{Final Answer
Correct Solutions}. This suggests that longer outputs may increase the likelihood of ``false positives''. Meanwhile, an important observation is that the average length of \textit{False Positive Solutions} in Omni-MATH100 is shorter than in AIME, while the corresponding false positive rate for Omni-MATH100 is significantly higher. This indicates that output length alone does not fully explain the trend; instead, the inherent difficulty of the benchmark plays a critical role in affecting the model’s susceptibility to ``false positives''.

\begin{figure}[t]
\centering  
{
\includegraphics[width=6cm,height = 4cm]{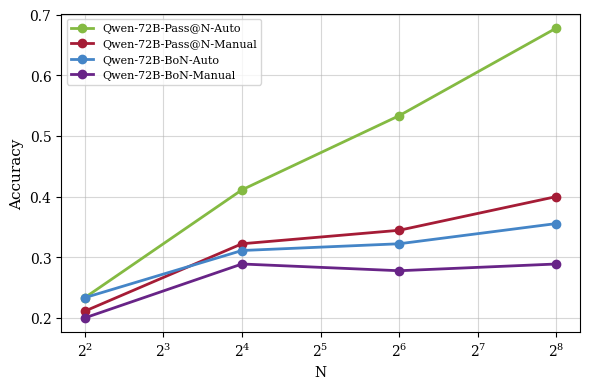}
}
\caption{\textbf{Pass@N vs. Best-of-N}. Qwen2.5-Math-72B-Instruct is utilized as the policy model, Qwen2.5-Math-RM-72B serves as the reward model, and AIME is employed as the test set.}
\label{pass at n}
\end{figure}

\subsubsection{Is Automatic Pass@N the Upper Bound of Inference Scaling?}
To further investigate the potential and possible upper limit of inference scaling, we examine the relationship between Pass@N and Best-of-N in this section.

\noindent\textbf{Pass@N yields significantly more correct answers than Best-of-N, but exhibits a substantially higher false positive rate.} Figure \ref{pass at n} demonstrates that while the gap between automatic Pass@N accuracy and automatic Best-of-N accuracy is substantial, the difference between manual Pass@N accuracy and manual Best-of-N accuracy is comparatively smaller. So in solution-level inference scaling methods, despite the use of an oracle reward model capable of detecting ``false positives'' across all responses, the inherent limitations of the policy model hinder the reward model's ability to select as many truly correct responses as expected. These limitations pose challenges to the broader application of inference scaling.

\subsubsection{How Does Rule-Based GRPO Affect False Positives?}
Recently, there has been a surge of research leveraging reinforcement learning to enhance mathematical reasoning, sparked by the release of DeepSeek-R1 \cite{guo2025deepseek}. Notable examples include SimpleRL-Zoo \cite{zeng2025simplerl}, DAPO \cite{yu2025dapo}, and Dr.GRPO \cite{liu2025understanding}. These studies show that reinforcement learning algorithms guided by rule-based reward functions—focus solely on final answer accuracy—can improve a model’s ability to reflect and verify, thereby strengthening its mathematical reasoning capabilities. In this section, we aim to examine the impact of rule-based GRPO on the occurrence of ``false positives''.

We compare the performance of Qwen2.5-Math-1.5B-Oat-Zero \cite{liu2025understanding}, a model trained using Dr.GRPO, with that of Qwen2.5-Math-1.5B-Instruct. The results are presented in Table \ref{tab:qwen_1.5b_comparison}. As shown, Qwen2.5-Math-1.5B-Oat-Zero exhibits a higher false positive rate and demonstrates limited reflection and self-verification. We attribute this to the use of a rule-based reward function that provides no supervision over intermediate reasoning steps, potentially contributing to the elevated false positive rate. However, in scenarios where GRPO effectively enhances self-reflection and verification, we hypothesize that the false positive rate may be reduced.

\begin{table}[htbp]
    \centering
    \small
    \caption{\textbf{Comparison Between Qwen2.5-Math-1.5B-Oat-Zero and Qwen2.5-Math-1.5B-Instruct}. Using AIME benchmark, we employ Qwen2.5-Math-RM-72B as the reward model and adopt the Best-of-N method.}
    \label{tab:qwen_1.5b_comparison}
    \resizebox{0.48\textwidth}{!}{
    \begin{tabular}{@{}ccc@{}}
        \toprule
        \textbf{Model} & \textbf{Best-of-256} & \textbf{False Positive Rate} \\
        \midrule
        \texttt{Qwen2.5-Math-1.5B-Oat-Zero} & 0.300 & 0.259 \\
        \texttt{Qwen2.5-Math-1.5B-Instruct} & 0.278 & 0.160 \\
        \bottomrule
    \end{tabular}
    }
\end{table}

\subsubsection{Analysis of False Positive Examples}
\label{quantitative analysis}
Through manual inspection of the model's outputs, we find that ``false positives'' primarily fall into the categories outlined in Section \ref{human detection method}. Some examples are in Appendix \ref{false positive examples}.

\begin{figure}[h]
\centering  
{
\includegraphics[width=6cm,height = 4cm]{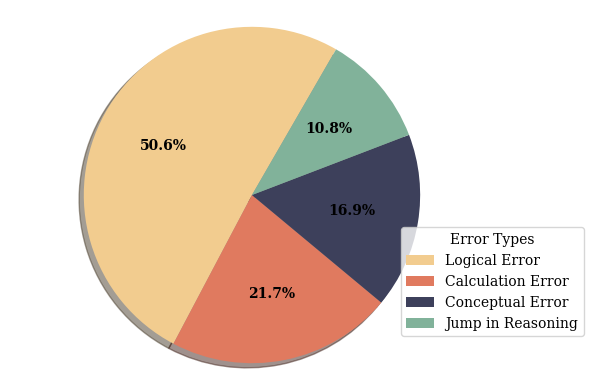}
}
\caption{\textbf{Error Type Statistics}. We count the false positive error types under the settings of Figure \ref{different model}, Figure \ref{different dataset}, and the $N=256$ configuration in Figure \ref{false positive}(b).}
\label{error type}
\end{figure}

We analyze and count the error types, with the results shown in Figure \ref{error type}. Notably, Logical Error constitutes the majority of ``false positives''. Moreover, We find that on relatively challenging datasets, general models exhibit a higher frequency of reasoning jumps and logical errors compared to mathematical models. This results in the general models false positive rate being significantly higher than that of the mathematical models on AIME benchmark in Figure \ref{different model}.

We also analyze the recent Long-CoT model. Specifically, we use DeepSeek-R1-Distill-Llama-70B \cite{guo2025deepseek} as the policy model and Qwen2.5-Math-RM-72B as the reward model on AIME. We adopt the Weighted Self-Consistency method with $N=64$ to perform false positive analysis. We observe that a large proportion of <answer> parts omit critical details of the solution process, making it difficult to verify the correctness of the reasoning steps. Furthermore, among the \textbf{69} solutions with correct final answers, we find \textbf{2} instances where the <answer> part contains intermediate reasoning errors, while their corresponding <think> parts are correct. This suggests a potential misalignment between <think> and <answer>.

\subsubsection{Deeper Analysis of False Positives Using t-SNE}
\label{deeper analysis}
We conduct a comprehensive analysis by organizing responses from LLaMA and Qwen model series into separate datasets, each containing questions, model responses, and labels (Correct, False Positive, and Incorrect). We analyze LLaMA responses using Llama-3.2-3B-Instruct and Llama-3.1-\{8,70\}B-Instruct, while Qwen responses are analyzed using Qwen2.5-Math-\{1.5, 7, 72\}B-Instruct. We then extract hidden states from the last 6 layers, apply either last-token or mean pooling strategies, and visualize the layer with the highest silhouette score using t-SNE dimensionality reduction, as shown in Figure \ref{fig:last_token-all_responses}, \ref{fig:mean-all_responses}. Our findings reveal that correct and incorrect responses tend to exhibit relatively distinct clustering patterns with observable boundaries. However, false positive responses do not form a separate, cohesive cluster. Instead, they are distributed across both correct and incorrect regions, with only a slight tendency to appear more frequently near incorrect responses than correct ones.

These results suggest that false positive responses lack distinctive representational features that would clearly differentiate them from either correct or incorrect responses in the model's internal representation space. This finding indicates that detecting ``false positives'' may be inherently challenging due to their intermediate nature between correct and incorrect responses.

\section{Discussions}

\textbf{Synthetic Mathematical Data.} Synthetic data has gained prominence due to the high cost associated with manual data creation. Current approaches often filter synthetic mathematical data based solely on the correctness of the final answer \cite{yu2023metamath, tong2024dart, luo2023wizardmath, liu2024acemath}. However, evaluating data quality exclusively based on the final answer can lead to ``false positives'', thereby introducing low-quality data into the dataset. To mitigate this issue, future research should prioritize the development of simple yet effective methods to accurately assess the correctness of intermediate reasoning steps, ensuring higher-quality synthetic data for training purposes.

\noindent\textbf{Self-Improvement in Mathematical Reasoning.} The current standard approach involves employing either solution-level sampling methods \cite{yuan2023scaling, chow2024inference} or tree search methods \cite{zhang2024rest, xie2024monte, shi2025efficient}, followed by filtering out correct answers using rule-based techniques. Subsequently, a reward model is utilized to score the model-generated responses, with high-quality outputs used for further training. However, as demonstrated in Section \ref{results}, it often generates a notable number of ``false positives'', which are also incorporated into subsequent training. These ``false positives'' are of lower quality, as they fail to provide meaningful insights into the reasoning steps, leading the model to primarily learn the answers rather than the underlying problem-solving logic. Consequently, the effectiveness of self-improvement may fall short of expectations.

\section{Conclusions}
In this paper, we have explored the ``false positive'' phenomenon in mathematical reasoning and investigated its impact on inference scaling. Through meticulous manual evaluation, we have demonstrated that ``false positives'' are prevalent across various models, datasets, and decoding methods. Furthermore, we have discussed the broader implications of ``false positives'' for critical applications such as self-improvement and data synthesis, where they can propagate errors and degrade the quality of generated outputs. This study underscores the need for more robust evaluation methods that prioritize both final-answer correctness and the validity of intermediate reasoning steps, offering a foundation for future research to enhance the accuracy and scalability of mathematical reasoning models.

\section*{Limitations}

While this study provides valuable insights into the phenomenon of ``false positives'' in the mathematical domain and their behavior under inference scaling, several limitations should be acknowledged. First, we do not conduct extensive experiments on the latest Long-CoT models. Second, our inference scaling tests are restricted to parallel sampling-based methods, without examining sequential revision-based approaches. Finally, due to resource constraints, we do not perform human detection on large-scale datasets. Despite these limitations, we believe this phenomenon is prevalent in the mathematical domain. Future research could expand the scope of datasets, explore additional inference scaling methods, and incorporate more Long-CoT models to validate the generalizability and robustness of these findings.

\bibliography{reference}

\clearpage

\appendix
\section{Implementation Details}
\label{implementation appendix}
\subsection{Model Detection Prompt}
\label{model detection prompt}
\begin{mycolorbox}[label=detection-prompt, breakable]{Prompt}{}
You are an expert mathematician and your task is to verify the correctness of a step-by-step solution to a math problem. Carefully analyze each step for logical consistency, mathematical accuracy, and adherence to any given formulas or rules. Disregard minor errors that do not affect the validity of the final answer or are irrelevant to it. \\

Problem: \\
\{problem\} \\

Solution: \\
\{solution\} \\

Based on the problem and solution provided above: \\
1. Output True if the solution is considered correct. \\
2. Output False if the solution is considered incorrect and contains some errors. \\

Please comprehensively evaluate all the steps in the solution and provide only True or False as your final output.
\end{mycolorbox}

\subsection{Solution-Level Inference Scaling Settings}
We employ the vLLM inference framework for our experiments. For the LLaMA series models, we set the sampling parameters to a temperature of 0.6 and a top\_p of 0.9, using the same prompts as in the official evaluation. Similarly, for the Qwen series models, we use a temperature of 0.7 and a top\_p of 0.8, maintaining consistency with the official evaluation settings. For DeepSeek-R1-Distill-Llama-70B, the parameters are set to a temperature of 0.6 and a top\_p of 0.95. For Qwen2.5-Math-1.5B-Oat-Zero, we use a temperature of 0.7 and a top\_p of 0.8. Due to the potential for negative outputs, Qwen2.5-Math-RM-72B can not directly apply the Weighted Self-Consistency method. To address this, we simply employ $\frac{reward-min}{max-min}$ as its final reward, where $reward$ represents the score of corresponding solution, $max$ and $min$ denote the maximum and minimum scores across $N$ model responses, respectively.

\subsection{Step-Level Inference Scaling Settings}
The vLLM inference framework is also employed for step-level inference scaling. Sampling parameters are set to a temperature of 0.7 and a top\_p value of 1.0 for both the LLaMA and Qwen series models. For the DVTS method, we configure the beam width to 4 and limit the process to a maximum of 40 iterations. In the MCTS method, we set the tree's maximum depth to 40 and the tree width to 4.

\section{False Positive Detection Benchmark}
\label{false positive detection benchmark}
We construct a false positive detection benchmark by leveraging both solution-level and step-level inference scaling methods. For solution-level inference scaling, we employ Llama-3.1-70B-Instruct and Qwen2.5-Math-72B-Instruct as policy models, evaluated on the AIME benchmark, with Qwen2.5-Math-RM-72B serving as the reward model. For step-level inference scaling, Llama-3.1-8B-Instruct and Qwen2.5-Math-7B-Instruct are utilized as policy models, assessed using the MATH100 benchmark, with Skywork-o1-Open-PRM-Qwen-2.5-7B as the reward model. Responses from Llama-3.1-8B-Instruct are selected using the Best-of-N method, while responses from Qwen2.5-Math-7B-Instruct are chosen via Weighted Self-Consistency. We adopt a sample size of $N=256$ for our analysis. Ultimately, this benchmark comprises \textbf{453} data points, designed to evaluate the detection capabilities of ``false positives'' across diverse inference scaling approaches.

\section{Algorithmic Details of Step-Level Inference Scaling}
\begin{breakablealgorithm}
    \caption{DVTS}
    \label{dvts}
\begin{algorithmic}
\STATE {\bfseries Input:} problem $p$, the number of candidate solutions $N$, beam width $M$, lookahead step $l$, the number of iterations $n$, policy model $\pi$, process reward model $V$
\STATE {\bfseries Output:} List of candidate solutions $L$

\STATE $num\_beams$ $\leftarrow N/M$
\STATE $Beam$ $\leftarrow$ Init($num\_beams$, $p$)

\FOR{$i=1$ {\bfseries to} $n$}
{
    \FOR{$j=1$ {\bfseries to} $num\_beams$}
    {
        \STATE Sample $M$ next steps from $\pi$
        \FOR{$k=1$ {\bfseries to} $M$}
        {
            \STATE Greedily generate $l$ steps based on corresponding history steps \textcolor{comment}{\COMMENT{Apply lookahead search}}
        }
        \ENDFOR
        \STATE Compute values using $V$ based on the next steps and lookahead steps
        \STATE Select the next step with highest value
    }
    \ENDFOR
    \STATE Update and prune $Beam$ \textcolor{comment}{\COMMENT{Remove completed beams}}
    \STATE Adjust $num\_beams$ \textcolor{comment}{\COMMENT{Decrement if beams terminate}}
    \STATE Append $M$ solutions to $L$ for each completed beam
}
\ENDFOR

\bfseries Return $L$
\end{algorithmic}
\end{breakablealgorithm}

\begin{breakablealgorithm}
    \caption{Vanilla MCTS}
    \label{mcts}
\begin{algorithmic}
\STATE {\bfseries Input:} problem $p$, the number of reasoning paths $n$, tree width $w$, tree max depth $d$, policy model $\pi$, process reward model $V$
\STATE {\bfseries Output:} List of trajectories $traj$

\STATE $T$ $\leftarrow$ Init($p$)\;
\STATE $traj$ $\leftarrow$ [ ]\;

\FOR{$i=1$ {\bfseries to} $n$}
{
    \STATE $traj\_single$ $\leftarrow$ $``"$\;
    \STATE $node$ $\leftarrow$ root($T$)\;
    \STATE $done$ $\leftarrow$ False\;
    \WHILE{\textbf{not} $done$}
    {
        \STATE $node$ $\leftarrow$ $\argmax\limits_{node'\in children(node)}v_{node'}+c\times P_{node'}\times\frac{\sqrt{N_{node}}}{N_{node'}+1}$ \textcolor{comment}{\COMMENT{Node Selection}}
        
        \STATE Update $traj\_single$ \textcolor{comment}{\COMMENT{Record action after active node update}}
        
        \STATE $done$ $\leftarrow$ terminated or truncated \textcolor{comment}{\COMMENT{terminated is controlled by eos\_token or sep\_token while truncated is determined by $d$}}
        
        \IF{\textbf{not} $done$ {\bfseries and} is\_leaf($node$)}
        {
            \STATE Expand($node$, $w$, $\pi$, $V$) \textcolor{comment}{\COMMENT{Expand the leaf node, get $w$ children, and assign them initial values}}
        }
        \ENDIF
    }
    \ENDWHILE
    \STATE Backpropagate($node$)
    \STATE Append $traj\_single$ to $traj$
}
\ENDFOR

\bfseries Return $traj$
\end{algorithmic}
\end{breakablealgorithm}

\onecolumn
\newpage

\section{False Positive Examples}
\label{false positive examples}

\begin{mycolorbox}[label=example-jump-in-reasoning, breakable]{Problem1}{}
    \textbf{Problem:} \\
    For any finite set $X$, let $| X |$ denote the number of elements in $X$. Define \[S_n = \sum | A \cap B | ,\] where the sum is taken over all ordered pairs $(A, B)$ such that $A$ and $B$ are subsets of $\left\{ 1 , 2 , 3,  \cdots , n \right\}$ with $|A| = |B|$. For example, $S_2 = 4$ because the sum is taken over the pairs of subsets \[(A, B) \in \left\{ (\emptyset, \emptyset) , ( \{1\} , \{1\} ), ( \{1\} , \{2\} ) , ( \{2\} , \{1\} ) , ( \{2\} , \{2\} ) , ( \{1 , 2\} , \{1 , 2\} ) \right\} ,\] giving $S_2 = 0 + 1 + 0 + 0 + 1 + 2 = 4$. Let $\frac{S_{2022}}{S_{2021}} = \frac{p}{q}$, where $p$ and $q$ are relatively prime positive integers. Find the remainder when $p + q$ is divided by 1000.
    \tcblower
    \textbf{Response:} \\
    \#\# Step 1: Understand the problem\\
    We are tasked with finding the value of $S_n$, which represents the sum of the sizes of the intersections of all pairs of subsets $(A, B)$ of the set $\{1, 2, 3, \cdots, n\}$ where $|A| = |B|$. We are then asked to find the ratio $\frac{S_{2022}}{S_{2021}}$ and express it as $\frac{p}{q}$, where $p$ and $q$ are relatively prime positive integers.
    
    \#\# Step 2: Break down the problem into cases\\
    To tackle this problem, let's first consider the possible cases for the intersection of two subsets $A$ and $B$. We can categorize these cases based on the size of the intersection, which can range from $0$ to $n$.
    
    \#\# Step 3: Calculate the number of pairs for each case\\
    For a given size $k$ of the intersection, we can calculate the number of pairs $(A, B)$ that satisfy this condition. This involves choosing $k$ common elements for the intersection and then selecting the remaining elements for $A$ and $B$ from the remaining $n-k$ elements.
    
    \#\# Step 4: Derive a formula for the sum of intersection sizes\\
    Using the calculated number of pairs for each case, we can derive a formula for the sum of the intersection sizes, $S_n$. This formula will involve a summation over all possible intersection sizes $k$.
    
    \#\# Step 5: Simplify the formula for $S_n$\\
    After deriving the formula, we can simplify it to obtain a more manageable expression for $S_n$. This simplification will help us in calculating the ratio $\frac{S_{2022}}{S_{2021}}$.
    
    \#\# Step 6: Calculate the ratio $\frac{S_{2022}}{S_{2021}}$\\
    Using the simplified formula for $S_n$, we can calculate the ratio $\frac{S_{2022}}{S_{2021}}$ and express it as $\frac{p}{q}$, where $p$ and $q$ are relatively prime positive integers.
    
    \#\# Step 7: Find the remainder when $p + q$ is divided by 1000\\
    Finally, we need to find the remainder when $p + q$ is divided by 1000 to obtain the final answer.
    
    The final answer is: $\boxed{245}$
\end{mycolorbox}
\noindent\begin{minipage}{\textwidth}
\captionof{figure}{\textbf{An Example of Jump in Reasoning}. The response lacks any intermediate calculations before providing the final answer.}
\end{minipage}

\begin{mycolorbox}[label=example-logical-error, breakable]{Problem2}{}
    \textbf{Problem:} \\
    Let $a,$ $b,$ $c$ be real numbers such that \[|ax^2 + bx + c| \le 1\]for all $0 \le x \le 1.$  Find the largest possible value of $|a| + |b| + |c|.$
    \tcblower
    \textbf{Response:} \\
    To find the largest possible value of \( |a| + |b| + |c| \) given that \( |ax^2 + bx + c| \le 1 \) for all \( 0 \le x \le 1 \), we start by examining the polynomial \( f(x) = ax^2 + bx + c \) at specific points in the interval \([0, 1]\). These points are \( x = 0 \), \( x = 1 \), and \( x = \frac{1}{2} \).

    First, we evaluate \( f(x) \) at these points:
    \[ f(0) = c, \]
    \[ f(1) = a + b + c, \]
    \[ f\left(\frac{1}{2}\right) = \frac{a}{4} + \frac{b}{2} + c. \]
    
    Since \( |f(x)| \le 1 \) for all \( 0 \le x \le 1 \), we have the following inequalities:
    \[ |c| \le 1, \]
    \[ |a + b + c| \le 1, \]
    \[ \left| \frac{a}{4} + \frac{b}{2} + c \right| \le 1. \]
    
    We want to maximize \( |a| + |b| + |c| \). To do this, we consider the system of equations formed by setting \( f(x) \) to its boundary values at these points. \textcolor{comment}{Let's assume \( f(0) = 1 \), \( f(1) = 1 \), and \( f\left(\frac{1}{2}\right) = -1 \).} This gives us the system:
    \[ c = 1, \]
    \[ a + b + c = 1, \]
    \[ \frac{a}{4} + \frac{b}{2} + c = -1. \]
    
    Substituting \( c = 1 \) into the other two equations, we get:
    \[ a + b + 1 = 1 \implies a + b = 0, \]
    \[ \frac{a}{4} + \frac{b}{2} + 1 = -1 \implies \frac{a}{4} + \frac{b}{2} = -2. \]
    
    Since \( a + b = 0 \), we have \( b = -a \). Substituting \( b = -a \) into the second equation, we get:
    \[ \frac{a}{4} + \frac{-a}{2} = -2 \implies \frac{a}{4} - \frac{2a}{4} = -2 \implies -\frac{a}{4} = -2 \implies a = 8. \]
    
    Then, \( b = -a = -8 \) and \( c = 1 \). Therefore, \( |a| + |b| + |c| = 8 + 8 + 1 = 17 \).
    
    To confirm that this is indeed the maximum value, we check if the polynomial \( f(x) = 8x^2 - 8x + 1 \) satisfies the condition \( |f(x)| \le 1 \) for all \( 0 \le x \le 1 \). The vertex of the parabola \( f(x) = 8x^2 - 8x + 1 \) is at \( x = -\frac{-8}{2 \cdot 8} = \frac{1}{2} \). Evaluating \( f(x) \) at the vertex, we get:
    \[ f\left(\frac{1}{2}\right) = 8\left(\frac{1}{2}\right)^2 - 8\left(\frac{1}{2}\right) + 1 = 8 \cdot \frac{1}{4} - 4 + 1 = 2 - 4 + 1 = -1. \]
    
    At the endpoints \( x = 0 \) and \( x = 1 \), we have:
    \[ f(0) = 1, \]
    \[ f(1) = 8 \cdot 1^2 - 8 \cdot 1 + 1 = 8 - 8 + 1 = 1. \]
    
    Since \( |f(x)| \le 1 \) for all \( 0 \le x \le 1 \), the value \( |a| + |b| + |c| = 17 \) is indeed the maximum possible value.
    
    Thus, the largest possible value of \( |a| + |b| + |c| \) is \(\boxed{17}\).
\end{mycolorbox}
\noindent\begin{minipage}{\textwidth}
\captionof{figure}{\textbf{An Example of Logical Error}. We can observe the reliance on unjustified assumptions: the model assumes \( f(0) = 1 \), \( f(1) = 1 \), and \( f\left(\frac{1}{2}\right) = -1 \) and then get the final answer, without explaining the rationality of this assumption.}
\end{minipage}

\begin{mycolorbox}[label=example-calculation-error, breakable]{Problem3}{}
    \textbf{Problem:} \\
    If \[\sqrt[3]{2} = a + \cfrac{1}{b + \cfrac{1}{c + \cfrac{1}{d + \dotsb}}},\]where $a,$ $b,$ $c,$ $d$ are positive integers, compute $b.$
    \tcblower
    \textbf{Response:} \\
    To find the value of \( b \) in the expression
    \[
    \sqrt[3]{2} = a + \cfrac{1}{b + \cfrac{1}{c + \cfrac{1}{d + \dotsb}}},
    \]
    we start by isolating the fractional part. Since \( \sqrt[3]{2} \approx 1.25992 \), we have \( a = 1 \). Therefore, we can rewrite the equation as
    \[
    \sqrt[3]{2} = 1 + \cfrac{1}{b + \cfrac{1}{c + \cfrac{1}{d + \dotsb}}}.
    \]
    Subtracting 1 from both sides, we get
    \[
    \sqrt[3]{2} - 1 = \cfrac{1}{b + \cfrac{1}{c + \cfrac{1}{d + \dotsb}}}.
    \]
    Next, we take the reciprocal of both sides:
    \[
    b + \cfrac{1}{c + \cfrac{1}{d + \dotsb}} = \frac{1}{\sqrt[3]{2} - 1}.
    \]
    To simplify \( \frac{1}{\sqrt[3]{2} - 1} \), we multiply the numerator and the denominator by the conjugate of the denominator, which is \( (\sqrt[3]{2})^2 + \sqrt[3]{2} + 1 \):
    \[
    \frac{1}{\sqrt[3]{2} - 1} \cdot \frac{(\sqrt[3]{2})^2 + \sqrt[3]{2} + 1}{(\sqrt[3]{2})^2 + \sqrt[3]{2} + 1} = \frac{(\sqrt[3]{2})^2 + \sqrt[3]{2} + 1}{(\sqrt[3]{2})^3 - 1^3} = \frac{(\sqrt[3]{2})^2 + \sqrt[3]{2} + 1}{2 - 1} = (\sqrt[3]{2})^2 + \sqrt[3]{2} + 1.
    \]
    Since \( (\sqrt[3]{2})^2 = \sqrt[3]{4} \), we have
    \[
    (\sqrt[3]{2})^2 + \sqrt[3]{2} + 1 = \sqrt[3]{4} + \sqrt[3]{2} + 1.
    \]
    \textcolor{comment}{We know that \( \sqrt[3]{4} + \sqrt[3]{2} + 1 \approx 3.31758 \), so the integer part is 3. Therefore, \( b = 3 \).}
    
    To confirm, we can check the next term. Subtracting 3 from \( \sqrt[3]{4} + \sqrt[3]{2} + 1 \), we get
    \[
    \sqrt[3]{4} + \sqrt[3]{2} + 1 - 3 = \sqrt[3]{4} + \sqrt[3]{2} - 2.
    \]
    Taking the reciprocal, we have
    \[
    \frac{1}{\sqrt[3]{4} + \sqrt[3]{2} - 2}.
    \]
    Using the same method as before, we multiply the numerator and the denominator by the conjugate \( (\sqrt[3]{4})^2 - (\sqrt[2]{2} \cdot 2) + 4 \), but for simplicity, we can see that the next term will be such that the process repeats similarly.
    
    Thus, the value of \( b \) is
    \[
    \boxed{3}.
    \]
\end{mycolorbox}
\noindent\begin{minipage}{\textwidth}
\captionof{figure}{\textbf{An Example of Calculation Error}. The model incorrectly calculates \( \sqrt[3]{4} + \sqrt[3]{2} + 1 \approx 3.31758 \).}
\end{minipage}

\begin{mycolorbox}[label=example-conceptual-error, breakable]{Problem4}{}
    \textbf{Problem:} \\
    The superfactorial $n\$$ is defined as $n\$ = \underbrace{ n!^{{n!}^{{\cdot}^{{\cdot}^{{\cdot}^{n!}}}}}}_{n!}$. What is the units digit of $4\$$?
    \tcblower
    \textbf{Response:} \\
    \textcolor{comment}{To determine the units digit of \(4\$\), we start by evaluating the expression \(4\$ = 4!^{4!^{4!^{4!}}}\).} First, we calculate \(4!\):
    \[
    4! = 4 \times 3 \times 2 \times 1 = 24.
    \]
    Thus, \(4\$ = 24^{24^{24^{24}}}\). Next, we need to find the units digit of \(24^{24^{24^{24}}}\). Notice that the units digit of a power of a number depends only on the units digit of the base. The units digit of \(24\) is \(4\), so we need to determine the units digit of \(4^{24^{24^{24}}}\).
    
    To do this, we observe the pattern in the units digits of the powers of \(4\):
    \[
    \begin{aligned}
    4^1 & = 4 \quad \text{(units digit is 4)}, \\
    4^2 & = 16 \quad \text{(units digit is 6)}, \\
    4^3 & = 64 \quad \text{(units digit is 4)}, \\
    4^4 & = 256 \quad \text{(units digit is 6)}, \\
    & \vdots
    \end{aligned}
    \]
    We see that the units digits repeat every 2 numbers: \(4, 6, 4, 6, \ldots\). Therefore, the units digit of \(4^n\) depends on whether \(n\) is odd or even. If \(n\) is odd, the units digit is \(4\), and if \(n\) is even, the units digit is \(6\).
    
    Since \(24^{24^{24}}\) is an even number (because any power of an even number is even), the units digit of \(4^{24^{24^{24}}}\) is \(6\).
    
    Thus, the units digit of \(4\$\) is \(\boxed{6}\).
\end{mycolorbox}
\noindent\begin{minipage}{\textwidth}
\captionof{figure}{\textbf{An Example of Conceptual Error}. The model misinterprets the meaning of \( n\$ \). Specifically, \( 4\$ \) should be represented as $\protect\underbrace{ 4!^{{4!}^{{\cdot}^{{\cdot}^{{\cdot}^{4!}}}}}}_{4!}$ rather than $\protect\underbrace{ 4!^{{4!}^{{\cdot}^{{\cdot}^{{\cdot}^{4!}}}}}}_{4}$}
\end{minipage}

\section{In-Depth Analysis of False Positives}

\begin{figure*}[t]
\centering
\subfigure[LLaMA-3B]
{
\includegraphics[width=7.75cm]{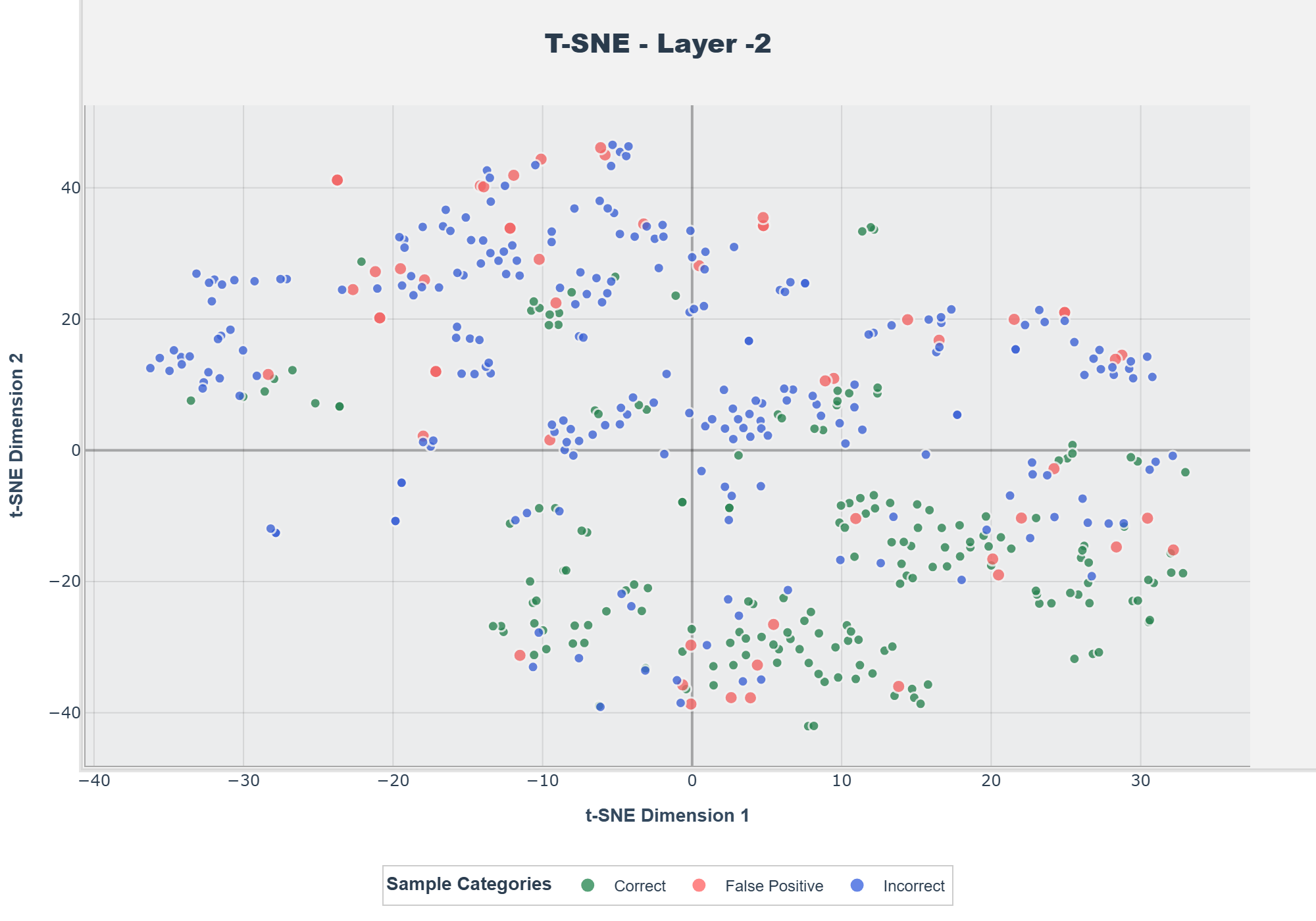}
}
\subfigure[LLaMA-8B]
{
\includegraphics[width=7.75cm]{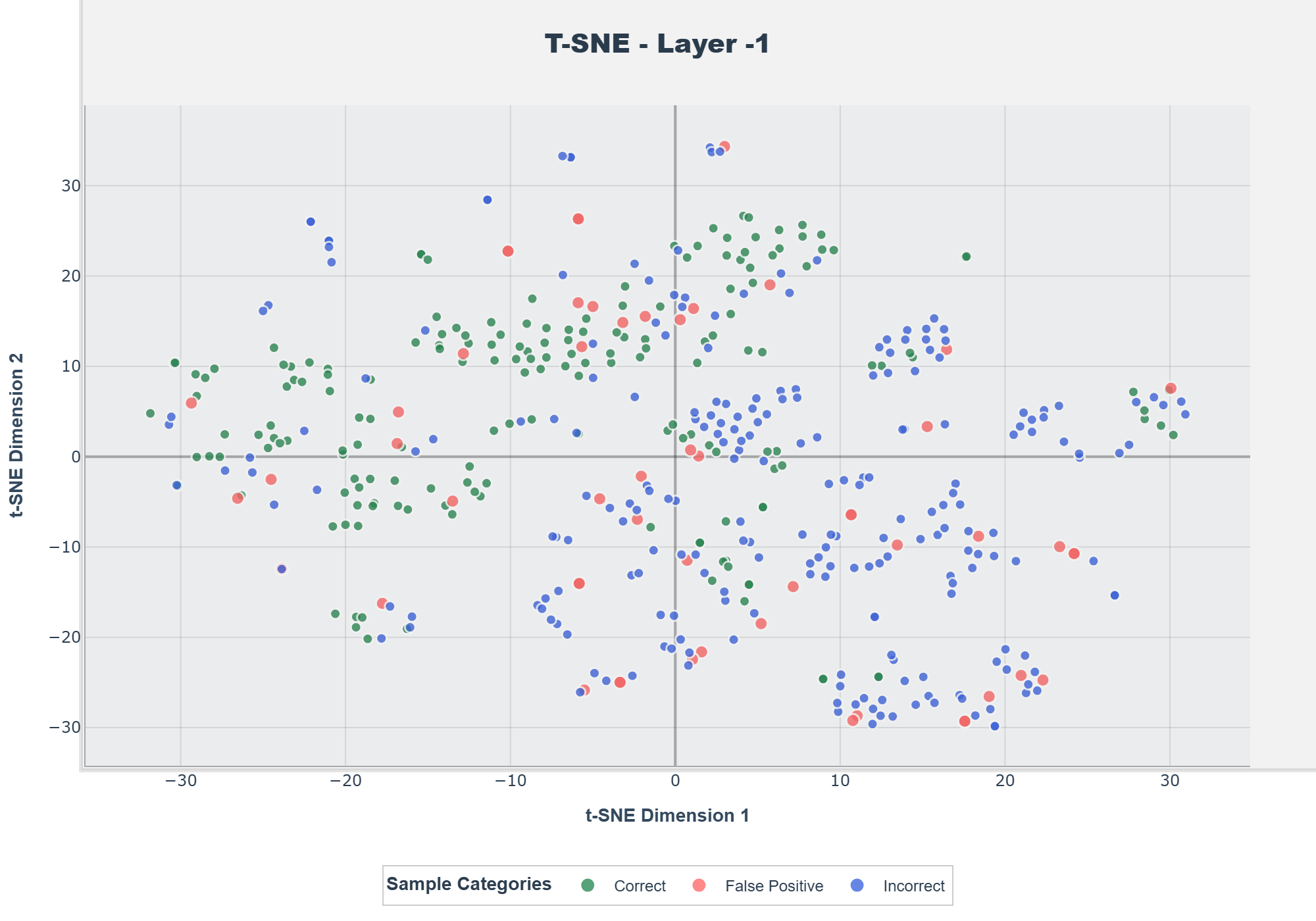}
}
\subfigure[LLaMA-70B]
{
\includegraphics[width=7.75cm]{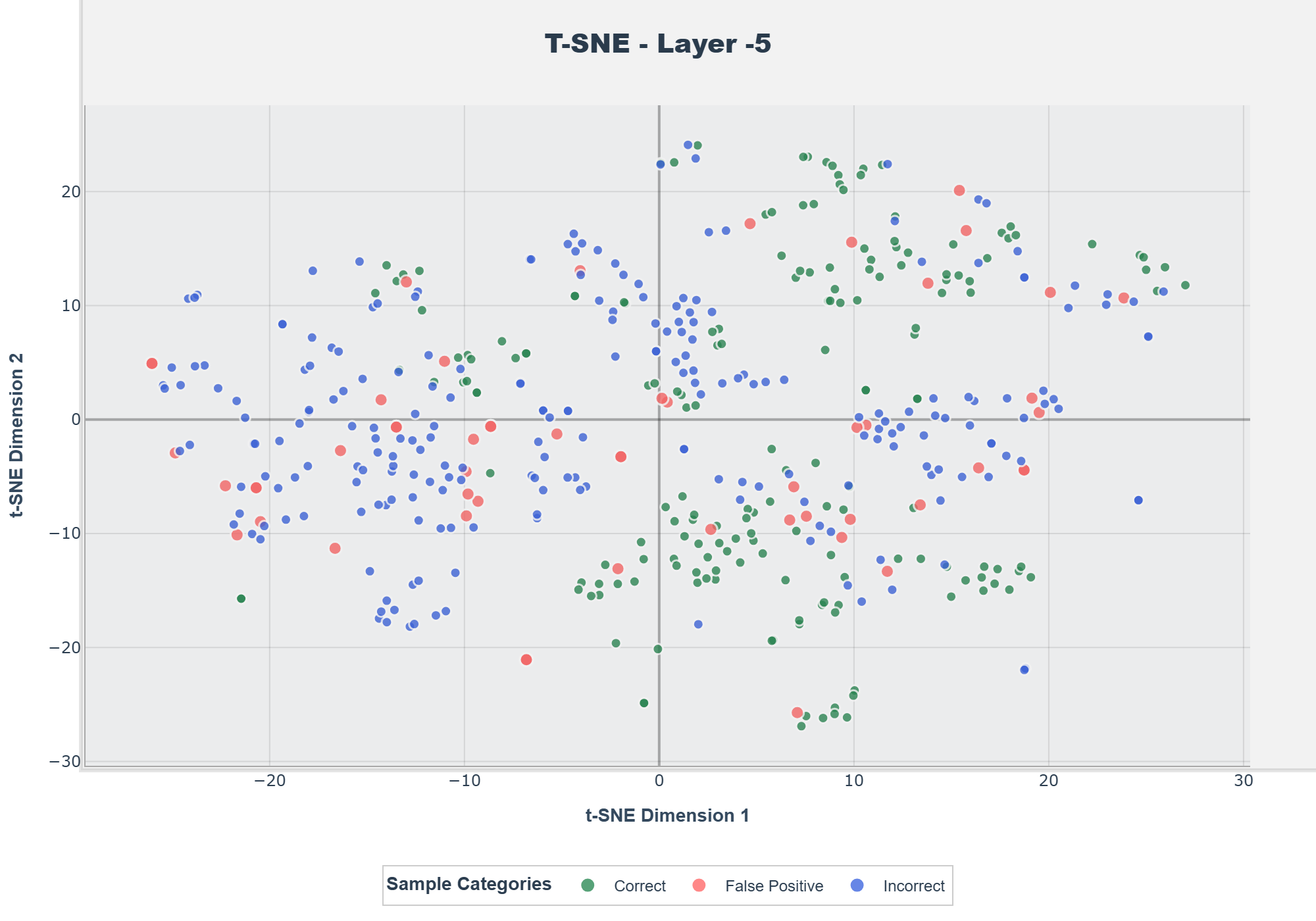}
}
\subfigure[Qwen-1.5B]
{
\includegraphics[width=7.75cm]{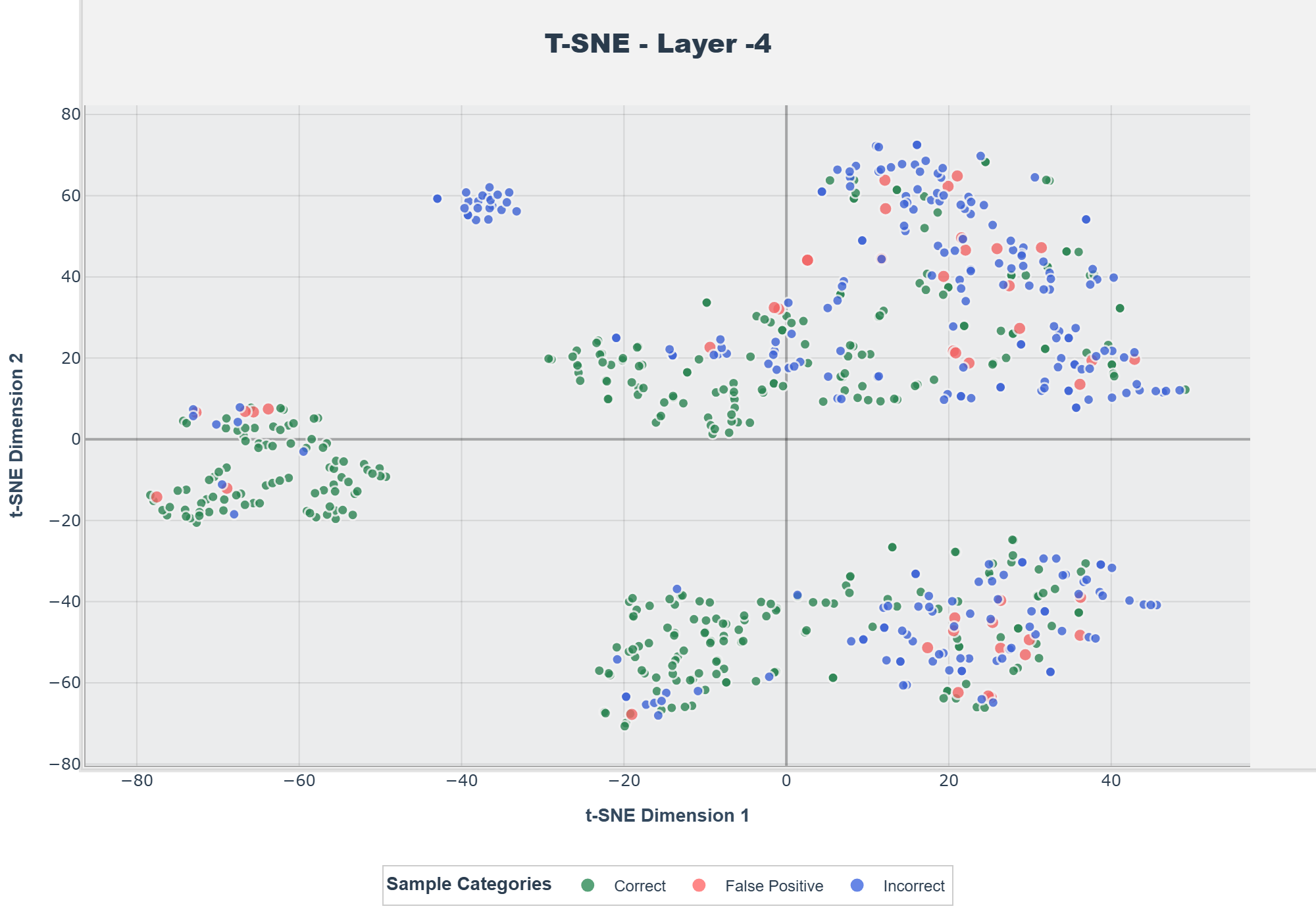}
}
\subfigure[Qwen-7B]
{
\includegraphics[width=7.75cm]{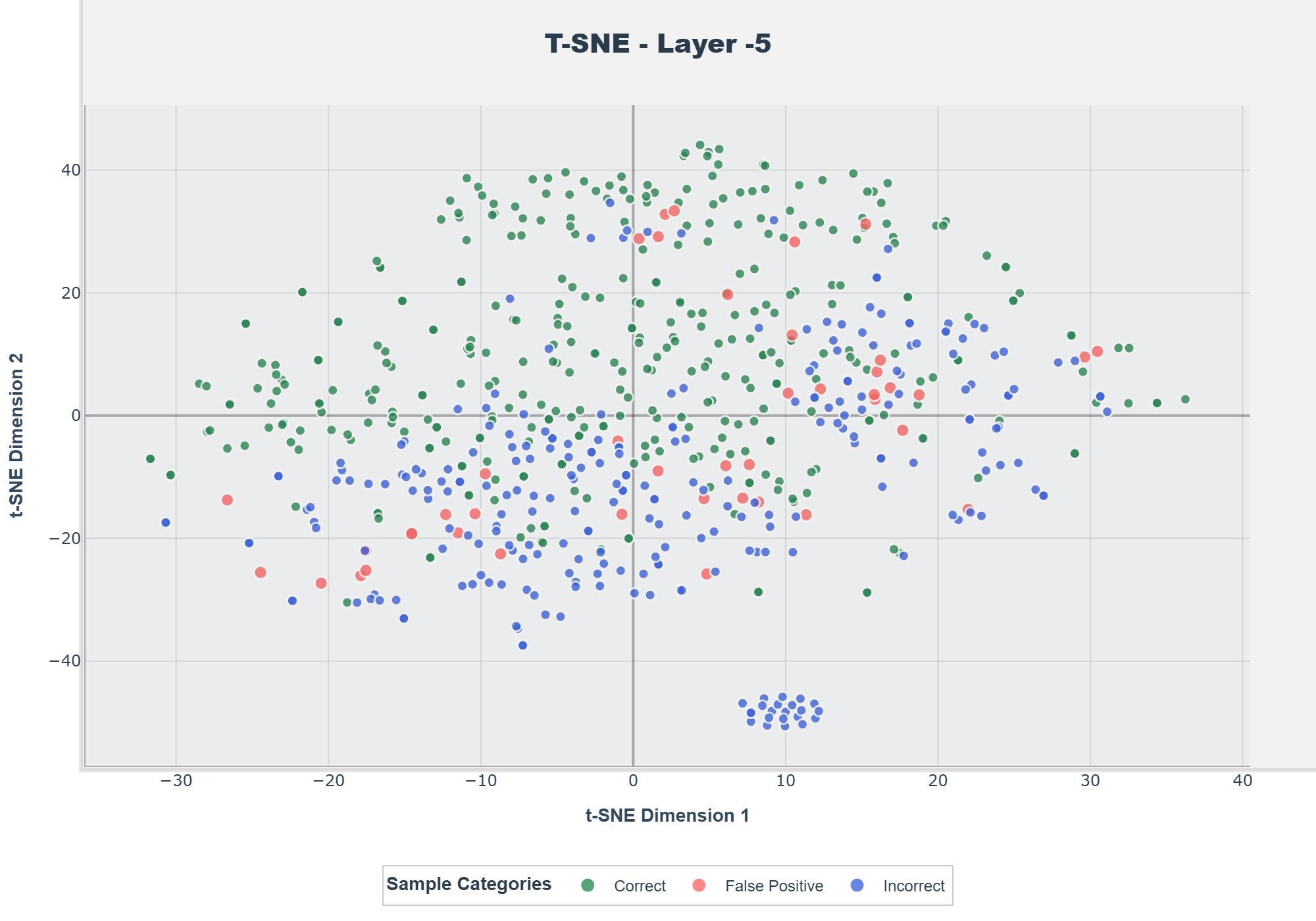}
}
\subfigure[Qwen-72B]
{
\includegraphics[width=7.75cm]{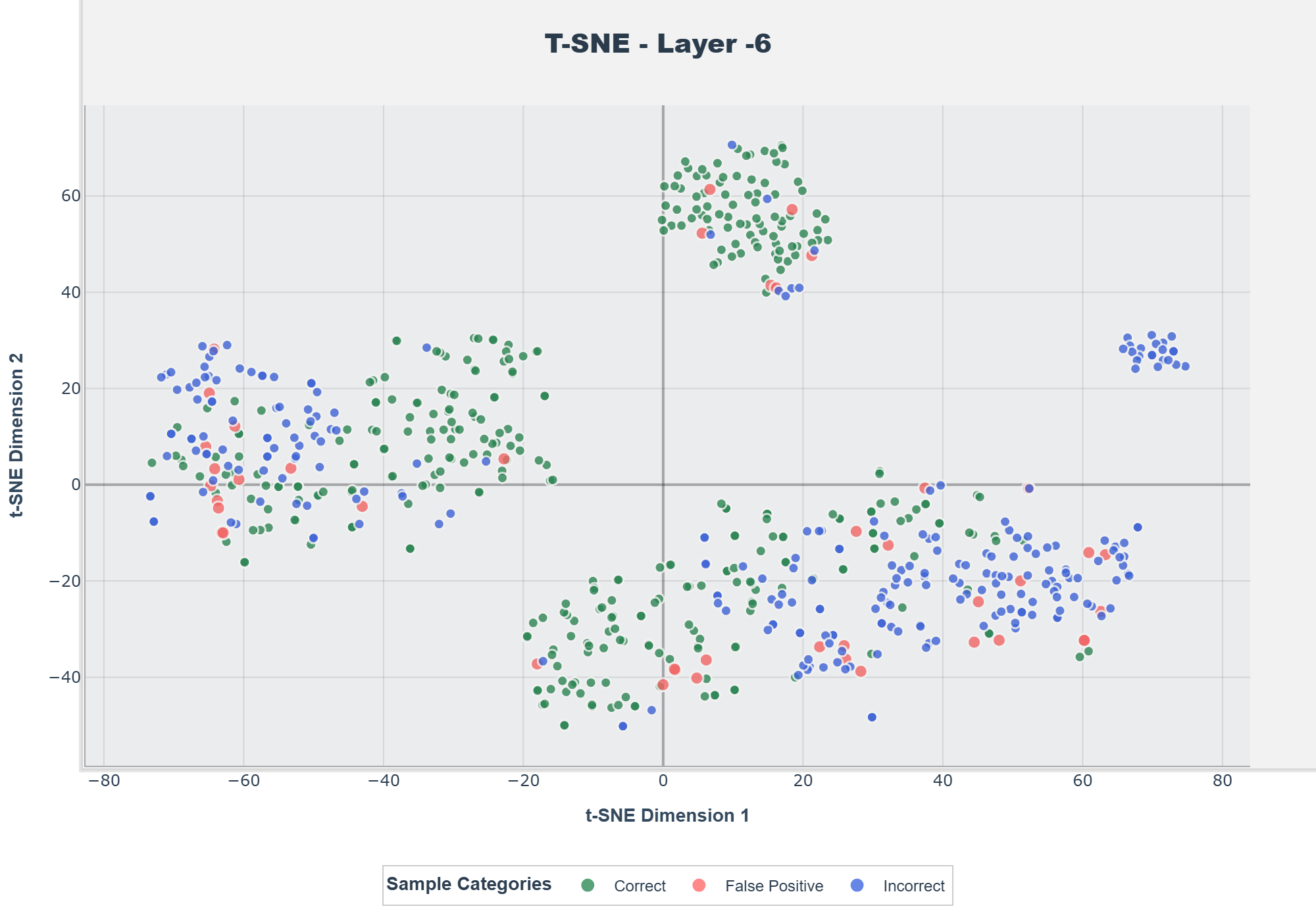}
}
\caption{\textbf{t-SNE Visualization with the Last Token Pooling Strategy}. We analyze LLaMA responses using Llama-3.2-3B-Instruct and Llama-3.1-\{8,70\}B-Instruct, while Qwen responses are analyzed using Qwen2.5-Math-\{1.5,7,72\}B-Instruct. Green, red, and blue dots indicate correct responses, ``false positives'', and incorrect responses, respectively.}
\label{fig:last_token-all_responses}
\end{figure*}

\begin{figure*}[b]
\centering
\subfigure[LLaMA-3B]
{
\includegraphics[width=7.75cm]{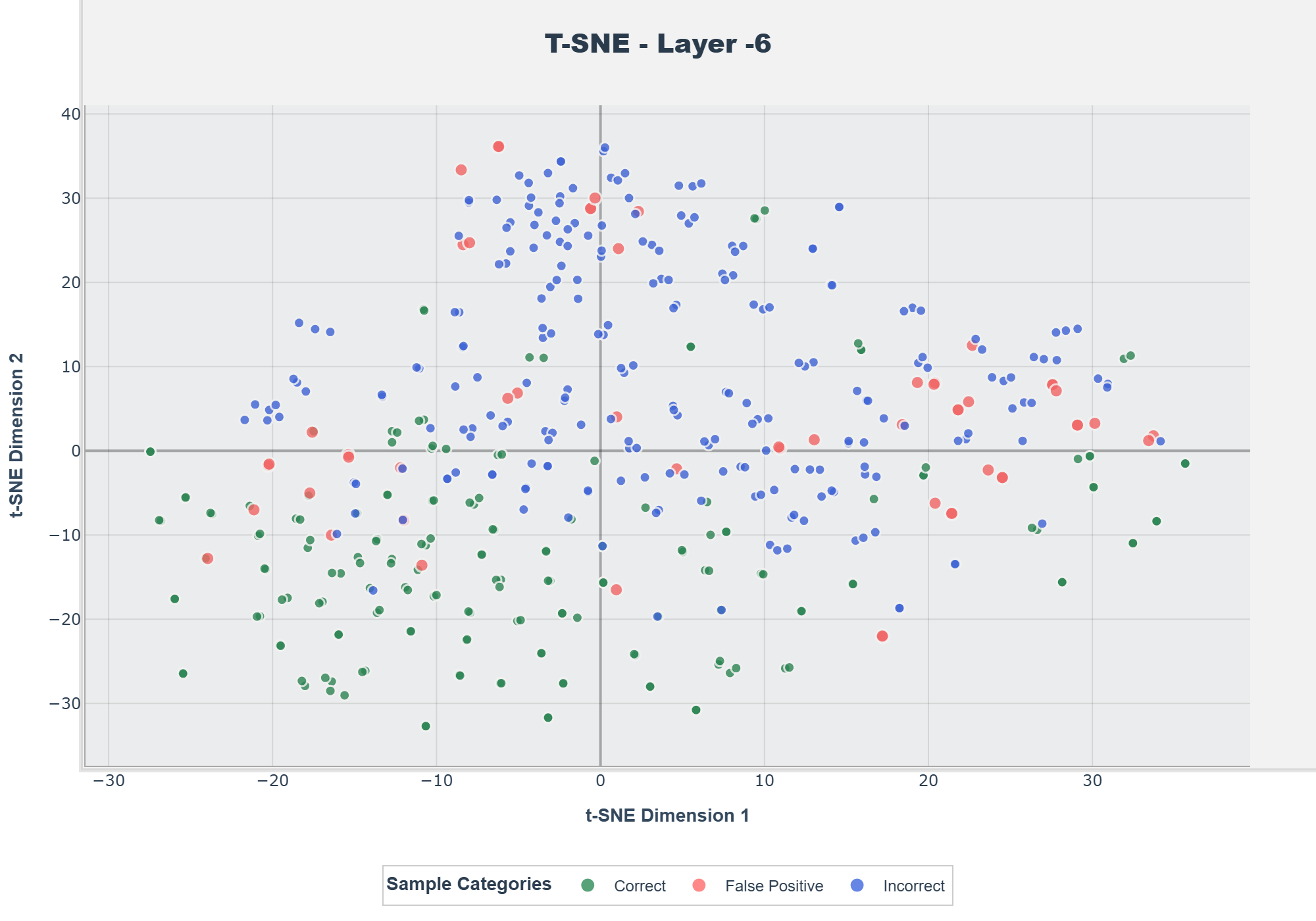}
}
\subfigure[LLaMA-8B]
{
\includegraphics[width=7.75cm]{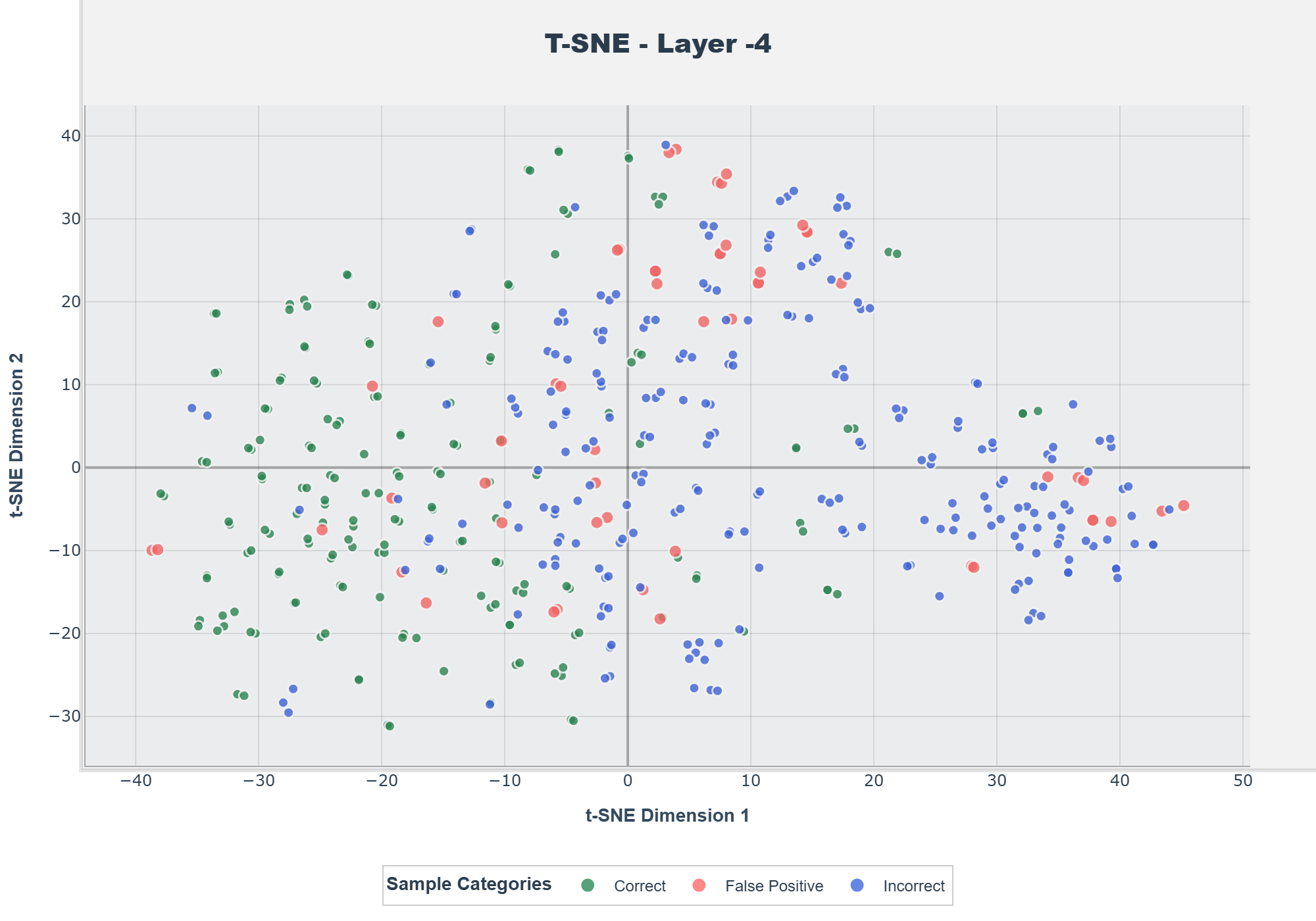}
}
\subfigure[LLaMA-70B]
{
\includegraphics[width=7.75cm]{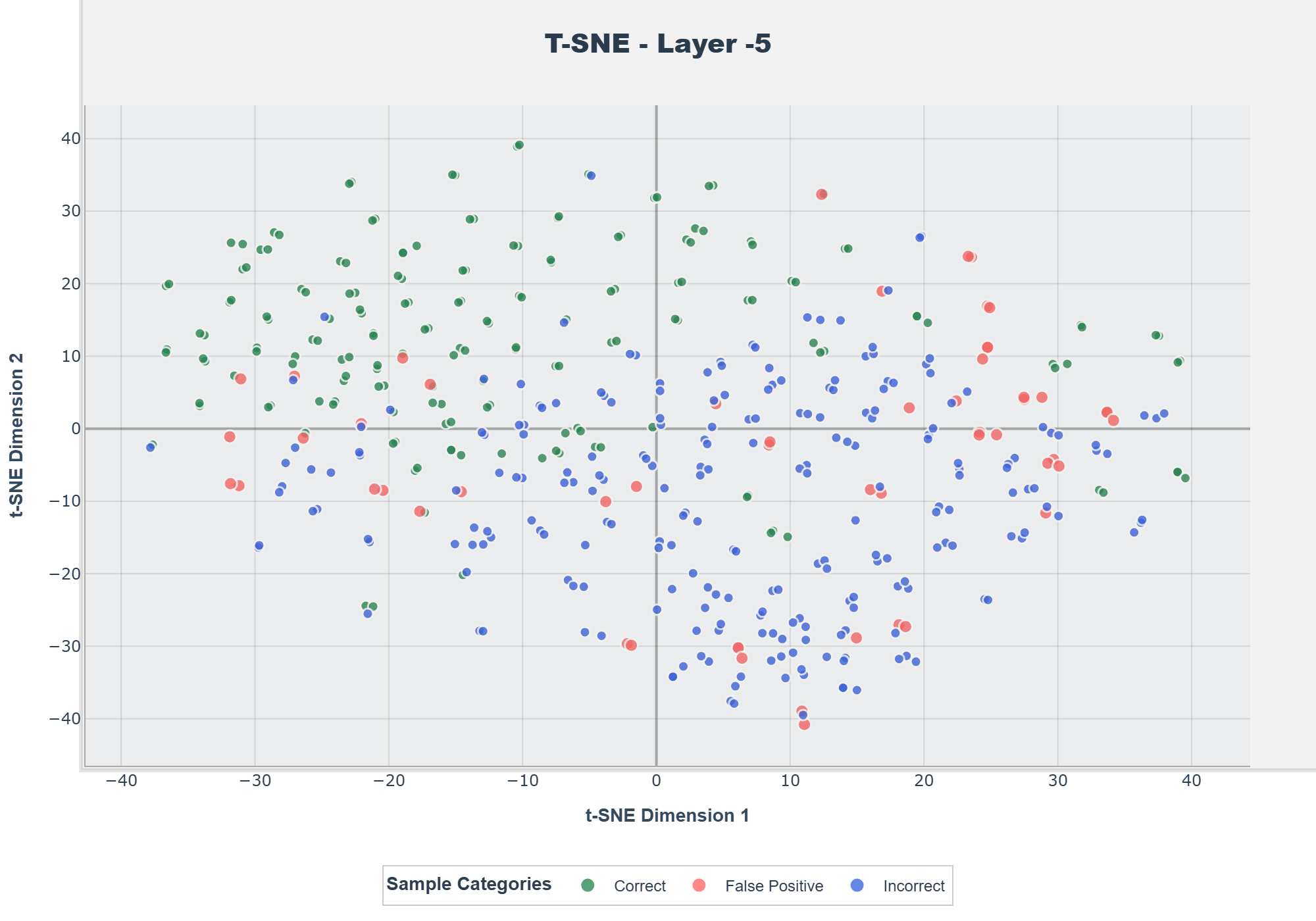}
}
\subfigure[Qwen-1.5B]
{
\includegraphics[width=7.75cm]{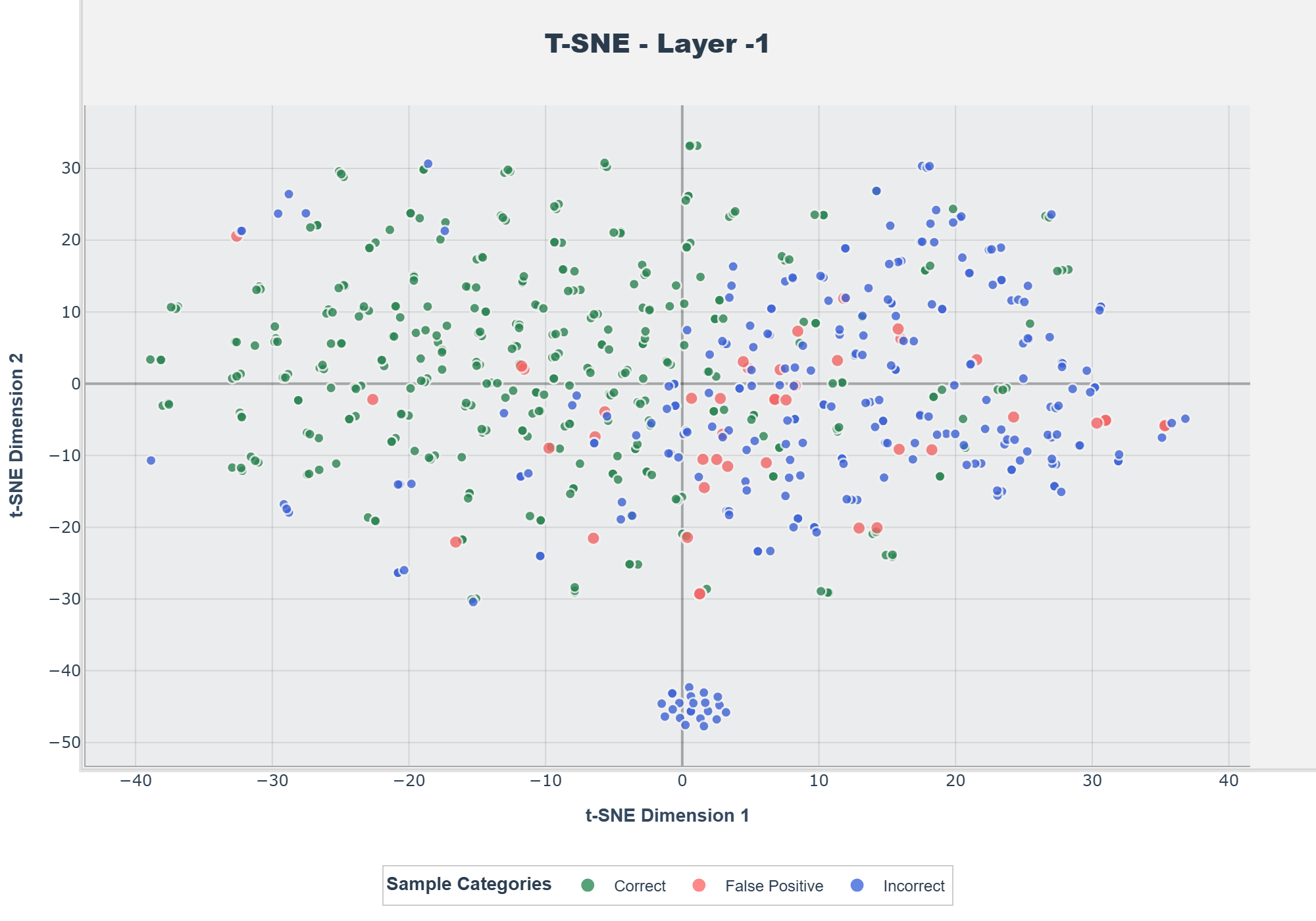}
}
\subfigure[Qwen-7B]
{
\includegraphics[width=7.75cm]{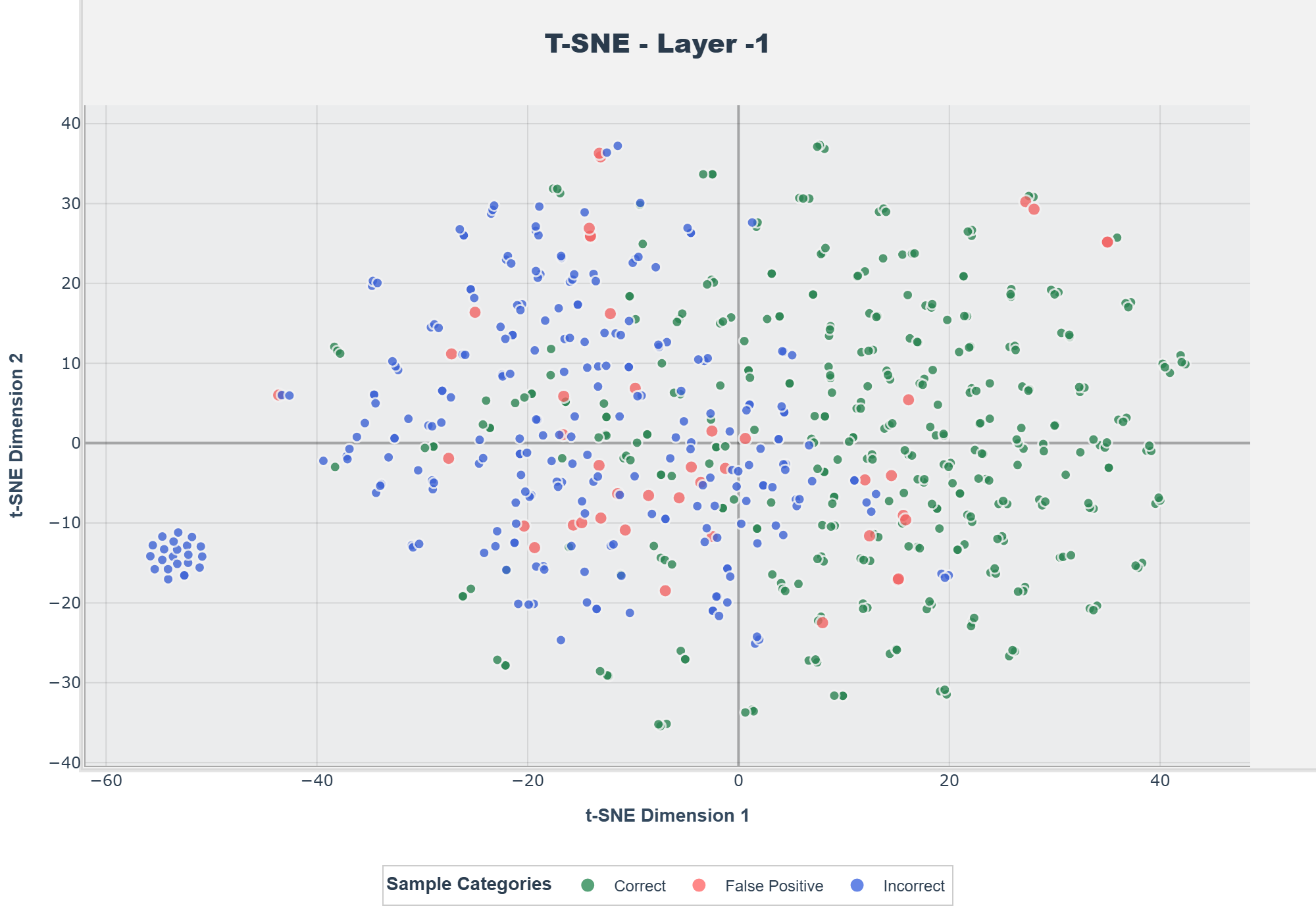}
}
\subfigure[Qwen-72B]
{
\includegraphics[width=7.75cm]{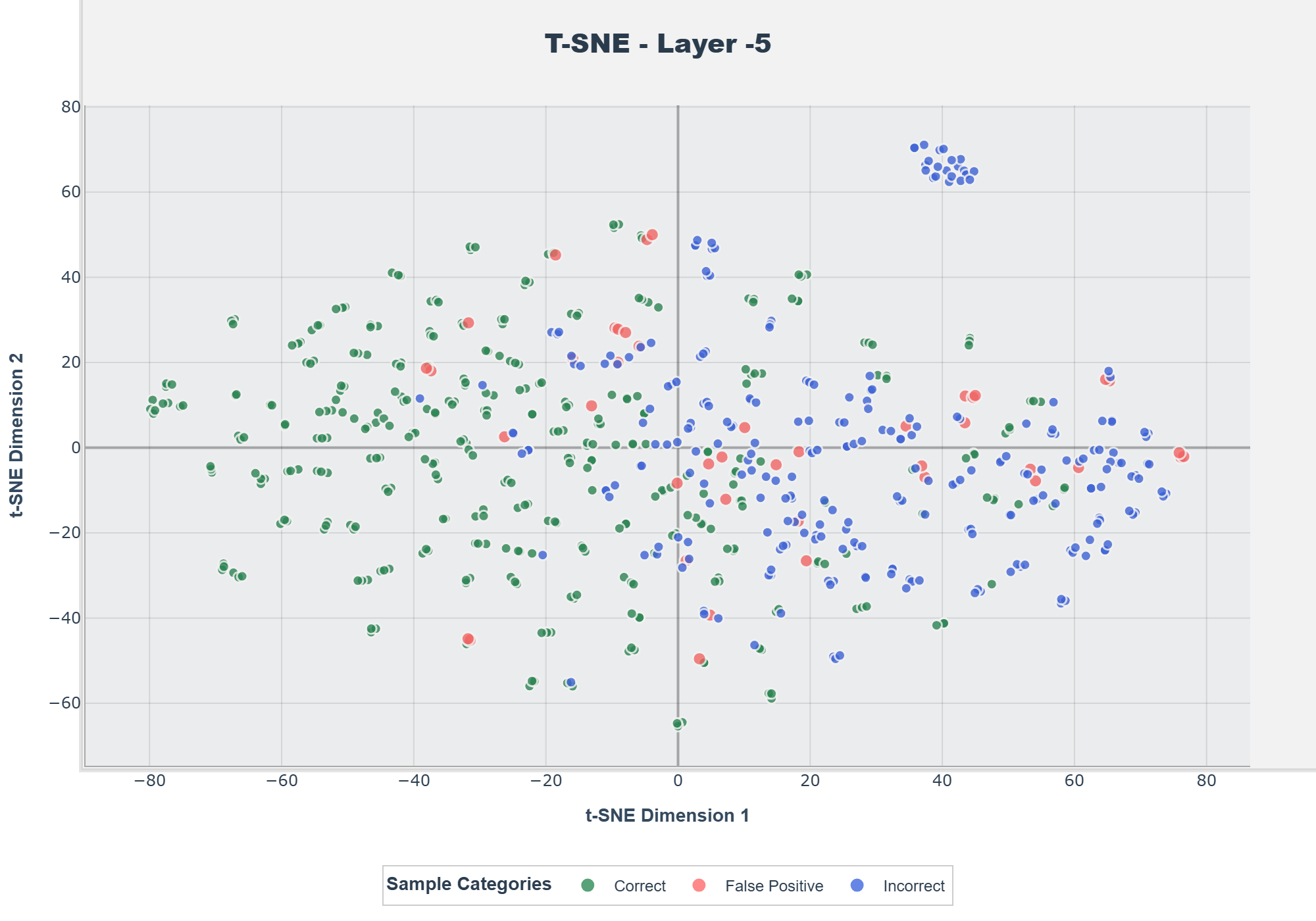}
}
\caption{\textbf{t-SNE Visualization with the Mean Pooling Strategy}. We analyze LLaMA responses using Llama-3.2-3B-Instruct and Llama-3.1-\{8,70\}B-Instruct, while Qwen responses are analyzed using Qwen2.5-Math-\{1.5,7,72\}B-Instruct. Green, red, and blue dots indicate correct responses, ``false positives'', and incorrect responses, respectively.}
\label{fig:mean-all_responses}
\end{figure*}

\end{document}